\documentclass[twocolumn]{svjour3}       
\smartqed  
\usepackage{graphicx}
\usepackage{moreverb,url}
\usepackage[bookmarksopen,bookmarksnumbered]{hyperref}
\usepackage{amsmath,amssymb,amsfonts,bm}
\usepackage{graphicx}
\usepackage{textcomp}
\usepackage{mathtools}
\usepackage{algpseudocode,algorithmicx,algorithm}  
\usepackage{url}
\usepackage{multirow}
\usepackage{epstopdf}
\usepackage{subfig}
\usepackage{caption}
\usepackage{comment}
\usepackage{makecell}
\usepackage{breakcites}
\usepackage{hyperref}

\newcommand\norm[1]{\left\lVert#1\right\rVert}
\DeclareMathOperator*{\argmin}{arg\,min}
\begin{document}

\title{Collaborative Localization for Micro Aerial Vehicles
}


\author{Sai Vemprala         \and
        Srikanth Saripalli 
}


\institute{S. Vemprala \at
			  Texas A\&M University, College Station, USA \\
			  (now at: Microsoft Corporation, Redmond, USA) \\
              \email{svemprala@tamu.edu}           
           \and
           S. Saripalli \at
              Texas A\&M University, College Station, USA \\
}

\date{Received: date / Accepted: date}

\maketitle

\begin{abstract}
  In this paper, we present a framework for performing collaborative localization for groups of micro aerial vehicles (MAV) that use vision based sensing. The vehicles are each assumed to be equipped with a forward-facing monocular camera, and to be capable of communicating with each other. This collaborative localization approach is developed as a decentralized algorithm and built in a distributed fashion where individual and relative pose estimation techniques are combined for the group to localize against surrounding environments. The MAVs initially detect and match salient features between each other to create a sparse reconstruction of the observed environment, which acts as a global map. Once a map is available, each MAV performs feature detection and tracking with a robust outlier rejection process to estimate its own pose in 6 degrees of freedom. Occasionally, one or more MAVs can be tasked to compute poses for another MAV through relative measurements, which is achieved by exploiting multiple view geometry concepts. These relative measurements are then fused with individual measurements in a consistent fashion. We present the results of the algorithm on image data from MAV flights both in simulation and real life, and discuss the advantages of collaborative localization in improving pose estimation accuracy.
\end{abstract}

\section{Introduction}
Micro aerial vehicles (MAV) are a special class of unmanned aerial vehicles that have gained both general attention and research focus in recent years. The term MAV is generally applied to small multirotor configurations: typically measuring $<$1m in any dimension, such as quadrotor platforms. MAVs possess several desirable properties: agile navigation in six degrees of freedom, a small size that allows them to fly within cluttered spaces, inexpensiveness and a relative ease of prototyping and so on. Currently, MAVs enjoy widespread popularity in many application domains: aerial photography, precision agriculture, search and rescue, delivery, inspection etc.

In the context of autonomous operations, MAVs require onboard sensing and computation for reliable localization and planning. The choice of sensors for MAV is typically constrained due to the requirement of small size: which results in a trade-off between fidelity of sensory information and size/power requirements. In this regard, vision sensors have shown great potential as exteroceptive sensors for MAVs. Specifically, monocular cameras can be seen as a particularly good fit for MAVs: as opposed to stereo cameras, which for instance, could possibly have baseline limitations and require more processing. Today, monocular cameras are almost ubiquitous on both hobby and research grade MAV platforms.

At the same time, the small size of MAVs usually creates constraints such as low computational power and smaller energy sources, which in turn limit their ability to perform complex tasks while maintaining sufficient flight time. Given these challenges, a single MAV in a complex autonomous operation can always run the risk of being resource limited with no backup in conditions that may lead to its failure. As a solution to this problem, it would be more desirable to employ multiple small, low-power MAVs as a team: an idea that can boost mission efficiency by allowing larger spatial coverage, larger distributed payloads etc. Multiple MAVs can also be leveraged for task distribution, helping reduce the computational burden on individual vehicles compared to single vehicle implementations. Collaboration can also help enhance localization accuracy as multiple sources of information can be fused for robust estimation. This is especially useful in the case of monocular vision sensing: while a single monocular camera cannot resolve the depth of a scene, depth can be computed using information from cameras on other vehicles in a group.

\begin{figure}
	\centering
	\subfloat[Collaboration between multiple MAVs to match feature points and to obtain a sparse reconstruction]{\includegraphics[width = 5.5cm,height=3cm]{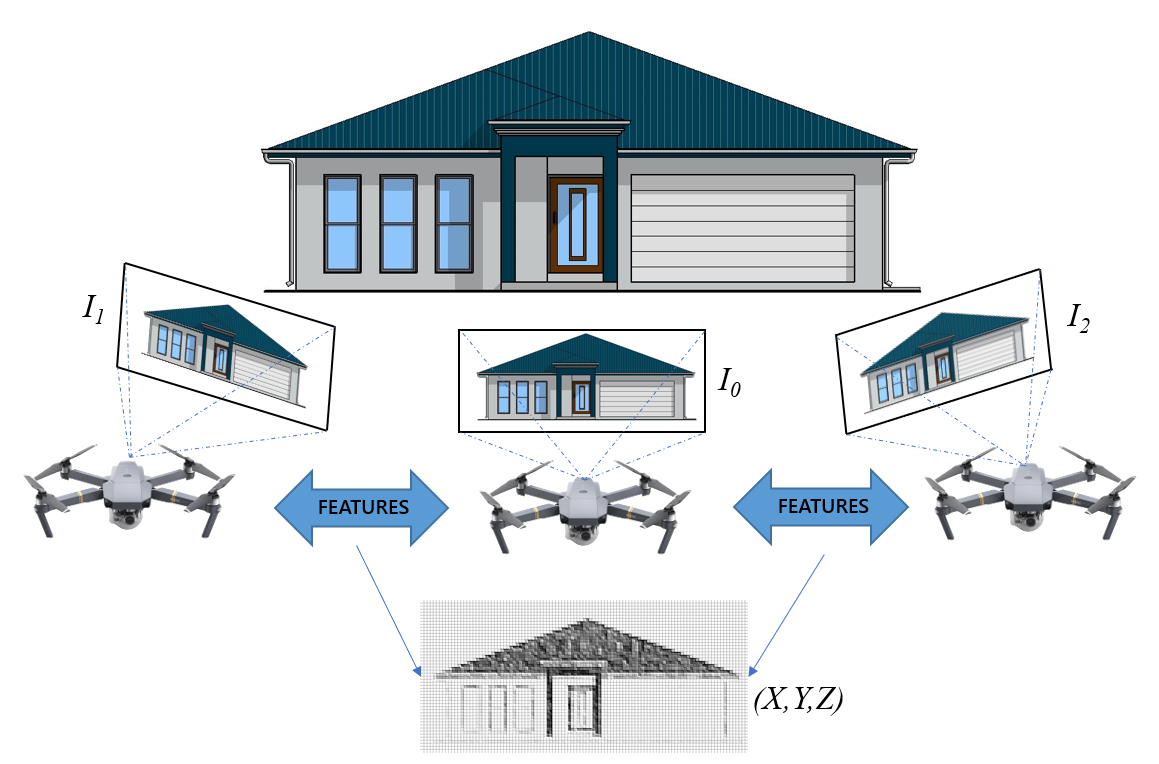}}
	\newline
	\subfloat[Two modes of localization: intra-MAV and inter-MAV]{\includegraphics[width = 5.5cm,height=3.5cm]{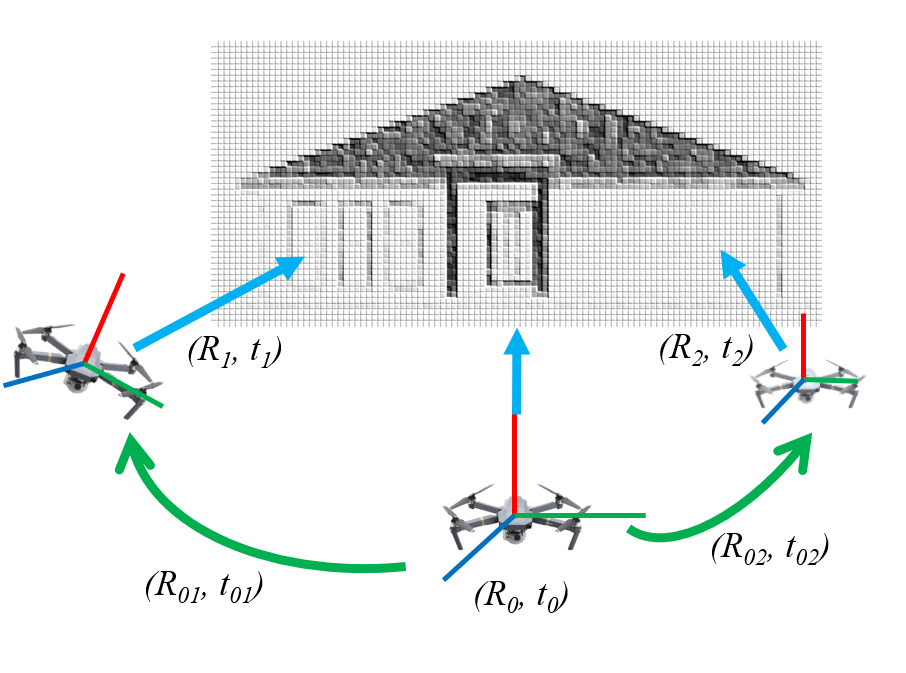}}
	\caption{Vision based Collaborative Localization : a conceptual representation}
	\label{fig:concept}
\end{figure}

In this paper, we present a framework for vision based collaborative localization (VCL) for a group of MAVs as an extension to our previous work \cite{8453412}, \cite{7759266}. We assume that each MAV is equipped with a forward-facing monocular camera, and is capable of communicating with the other MAVs to transmit or receive information. In the first step of the algorithm, the MAVs capture images of the environment visible through their cameras, upon which feature detection and matching are performed to isolate common salient features. These common features are then triangulated to form a sparse reconstruction that acts as a global map: which is then shared to all the vehicles (Figure \ref{fig:concept}(a)). Once the MAVs start moving, each MAV performs feature tracking to observe which features from the map are still visible, and uses these 2D-3D correspondences to perform its own individual pose estimation, which we call intra-MAV localization. When required, one or more MAVs can generate relative pose measurements to a target MAV and these estimates can be fused with the target's own individual estimate in a consistent way: and this process is known as inter-MAV localization (Figure \ref{fig:concept}(b)). As the MAVs continue to navigate, if the number of tracked features for the MAVs consistently falls below a threshold, the MAVs can match features between themselves to update the global map. Between the intra-MAV and inter-MAV modes of operation, the VCL algorithm operates in a mostly decentralized way. While communication is required between the members of the group, there is no need for it to be continuous or fully synchronous, and the algorithm has been designed in a way such that bandwidth requirements are reduced. In the next few sections, we present the details of our algorithm along with results from realistic simulation imagery and data from real MAV flights. Although the system has been tested offline on pre-recorded image datasets, we reserve some discussion for the applicability of the algorithm to real-time deployment.

\section{Related Work}

Vision based localization has been studied extensively in the literature. Initially, it was achieved through external camera placement such as in professional motion capture systems \cite{5196224} and \cite{michael2010grasp}. When vision sensors were used as onboard exteroceptive sensors, RGBD sensors were one of the initially investigated setups. Microsoft Kinect sensors were used for altitude estimation \cite{stowers2011altitude}, in tandem with a 2D laser rangefinder for mapping and localization \cite{shen2012autonomous} and visual odometry \cite{bachrach2012estimation}. Even more recently, full six degree-of-freedom localization was demonstrated using RGBD sensors \cite{fang2015real}. 

The ubiquity and compactness of monocular cameras have had a significant influence on their popularity and applicability for estimation in both computer vision and robotics communities. Many monocular camera based localization and mapping methods have been developed over the last decade such as PTAM \cite{klein2007parallel}, SVO \cite{forster2014svo}, ORB-SLAM2 \cite{mur2017orb} and LSD-SLAM \cite{engel2014lsd}, among which some were successfully implemented on UAV platforms. When applying these techniques onboard MAVs, a specific focus lies on removing the scale ambiguity. Various algorithms were proposed in the last few years that try to remove scale ambiguity either by fusing vision data with an IMU \cite{jones2011visual}, using ultrasonic rangefinders in conjunction with optical flow such as in the commercial autopilot PIXHAWK \cite{meier2012pixhawk}, and most recently, an algorithm was developed that can estimate depth from monocular imagery in a probabilistic fashion, but is computationally intensive \cite{pizzoli2014remode}. Many other promising monocular visual-inertial systems have been proposed, such as MSCKF \cite{mourikis2007multi}, visual-inertial ORB-SLAM2 \cite{mur2017visual}, VINS-MONO \cite{qin2018vins} etc.

Collaborative localization has been of interest over the past decade as well, with the theoretical foundations having been studied extensively. The general idea of collaborative localization has involved the concept of fusing different measurements to result in a more accurate fused state. Martinelli et al \cite{martinelli2005multi} present a localization approach that uses an extended Kalman filter to fuse proprioceptive and exteroceptive measurements, applied to multi-robot localization. Nerurkar et al \cite{nerurkar2009distributed} present a distributed cooperative localization algorithm through maximum aposteriori estimation, under the condition that continuous synchronous communication exists within a robot group. Carrillo-Arce et al \cite{carrillo2013decentralized} present a decentralized cooperative localization approach where robots need to communicate only during the presence of relative measurements, which was tested in simulation and on Pioneer ground robots. We use this algorithm in our paper to facilitate inter-MAV data fusion. Indelman et al \cite{indelman2014multi} propose a multi robot localization algorithm that can handle unknown initial poses and solves the data association problem through expectation maximization. Knuth and Barooah \cite{knuth2009distributed} propose a distributed algorithm for GPS-denied scenarios, where the robots fuse each other's information and average the relative pose data in order to achieve cooperative estimation.

Recently, collaborative localization ideas are being fused into the realm of aerial vehicles and vision based localization. Faigl et al \cite{faigl2013low} proposed a method involving onboard cameras and observation of black and white markers for relative localization within a MAV swarm. Indelman et al \cite{indelman2012distributed} propose a technique for cooperative localization for camera-equipped vehicles inspired by multi-view geometry ideas such as the trifocal tensor that estimates transformation between images. Zou and Tan \cite{zou2013coslam} present a collaborative monocular SLAM system with a focus on handling dynamic environments, with multiple vehicles helping each other isolate moving features from constant ones, but requiring constant communication between cameras. In \cite{achtelik2011collaborative}, the authors present an approach where two UAVs equipped with monocular cameras and IMUs estimate relative poses along with absolute scale, thus acting as a collaborative stereo camera. Piasco et al \cite{piasco2016collaborative} also present a distributed stereo system with multiple UAVs for collaborative localization with a focus on formation control. Forster et al \cite{forster2013collaborative} show a structure-from-motion based collaborative SLAM system, which contains a centralized ground station whose function is to merge maps created by various vehicles. In this framework, the vehicles do not benefit from additional information from other vehicles. Similarly, Schmuck and Chli \cite{schmuck2017multi} present a collaborative monocular SLAM pipeline for MAVs where each MAV runs the ORB-SLAM2 algorithm for SLAM and a central server focuses on place recognition, optimization and map fusion.

In our paper, we propose a vision-only based collaborative approach focused on localization of a group of MAVs. The main contribution of this work lies in the extension of vision based 6-DoF pose estimation for MAVs to a cooperative estimation scheme by combining individual and relative estimation. In addition, our relative localization approach does not require direct visual recognition or explicit range measurements between vehicles, and instead, this information is inferred through feature overlap. We list out the main features of our proposed approach, some of which exhibit differences compared to other existing works in this domain.

\begin{enumerate}
	\item In our approach, we formulate the problem as localization and not SLAM, emphasizing direct communication of relative measurements and fusion of relative and individual pose estimates. MAVs can readily receive corrections from other members in the group to enhance pose accuracy. Our approach is more suited for applications with task spaces that can allow high visual overlap, such as cooperative assembly or aerial imaging, where fusing information from multiple sources can benefit localization.
	\item Our approach is a vision-only approach with no requirement of alternative sensors for collaboration. Both individual and relative pose estimation are performed using only visual (feature) data. These estimates are also fused in a consistent manner, allowing the MAVs to benefit from other sources of information accurately. 
	\item The proposed algorithm is aimed to be distributed, as pose estimation is attempted individually by all MAVs. Communication between vehicles is only required when relative measurements are requested, either as part of pose correction or map construction. This, coupled with a reduced data representation using only feature information, helps reduce network/bandwidth requirements during constrained operation.
\end{enumerate} 

\section{Problem Statement}

\begin{figure*}
	\centering
	\includegraphics[width = 12cm,height=7cm]{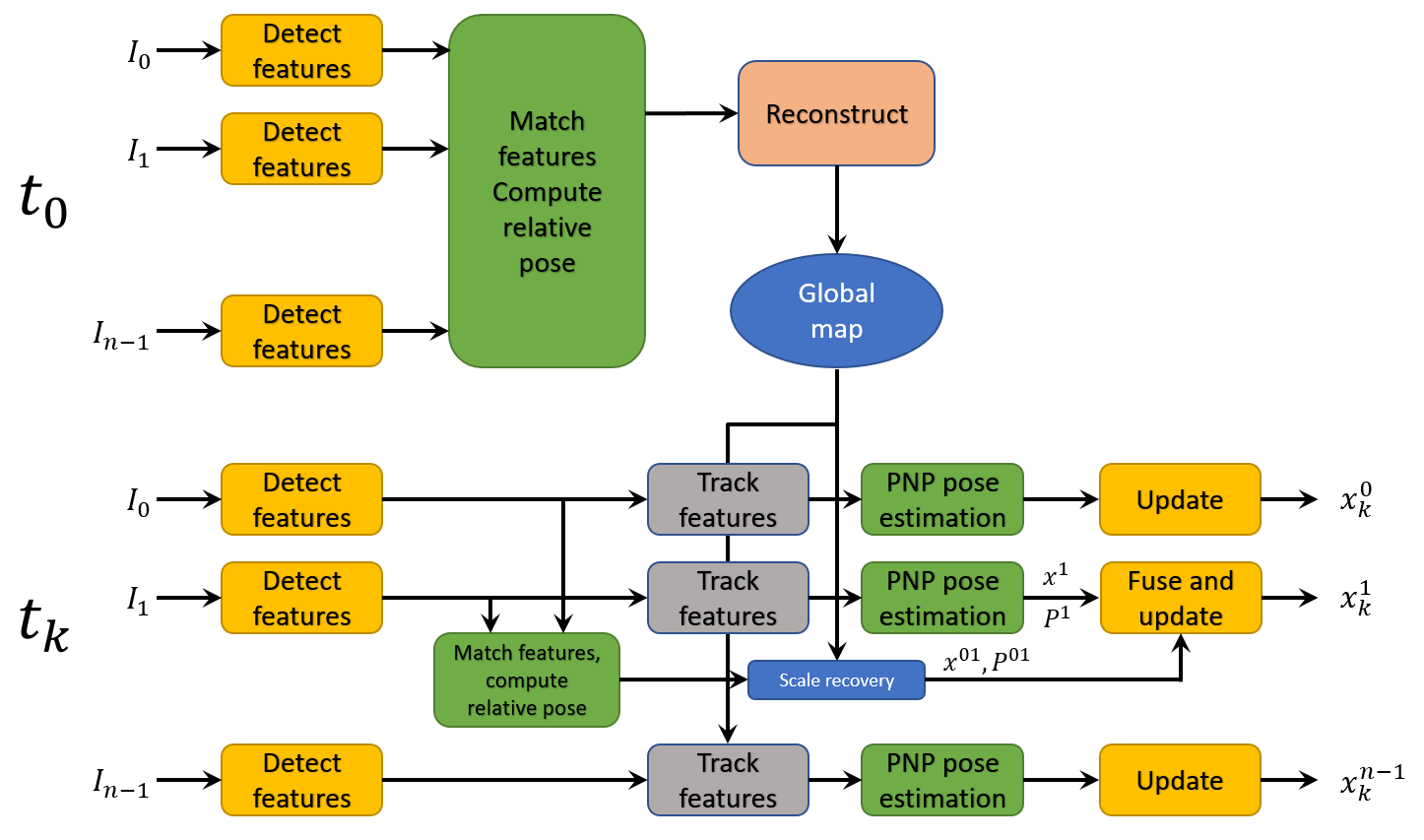} 
	\caption{Flow of the vision based collaborative localization (VCL) algorithm. At any time step $k$, an MAV can fuse estimates from individual (intra-MAV) localization and relative (inter-MAV) localization to produce a final pose estimate.}
	\label{fig:vcl}
\end{figure*}

The main goal of the proposed framework is to estimate six degree-of-freedom poses of multiple micro aerial vehicles (MAV). Each micro aerial vehicle is a multirotor platform capable of moving in all three translational axes with roll, pitch and yaw capabilities, thus navigating in $\mathbb{R}^3 \times SO(3)$, and is equipped with an intrinsically calibrated monocular camera. In the world frame of reference, as the vehicles are usually small in size, we consider the position of the camera to be equivalent to the position of the vehicle. Thus, for $k$ MAVs in a group, the goal is to estimate a system state $\mathbf{X}$, comprising of individual states: where each state is the 3D position of the  respective MAV's onboard camera and the roll, pitch, yaw angles, all in a predefined frame of reference. The reference frame can be chosen according to the application: it can represent a specific location in the world frame, or all the vehicles in a group can be made to localize relative to a leader MAV. As the problem is formulated as localization as opposed to full SLAM, the system state does not account for map points or any uncertainties associated with them. 

\begin{gather}
  \begin{align}
\mathbf{X} &= \begin{bmatrix} \mathbf{X_0} & \mathbf{X_1} & ... & \mathbf{X_n} \end{bmatrix} \\
\mathbf{X_m} &= \begin{bmatrix} x_m & y_m & z_m & \phi_m & \theta_m & \psi_m \end{bmatrix} 
\end{align}
\end{gather}

The cameras are assumed to adhere to the central projection model; each of which has a known intrinsics matrix $\mathbf{K}$ and is capable of producing 2D projections for each 3D point it observes through a projective mapping $\pi:\mathbb{R}^3 \implies \mathbb{R}^2$. At every time step, we assume the availability of images from each camera from which salient feature points are observed as 2D projections. Given this information, the responsibility of the algorithm is to estimate the 6-DoF poses of all MAVs to form the state matrix specified above.

To facilitate pose estimation, we make three important assumptions:
\begin{enumerate}
	\item All the cameras onboard the vehicles are calibrated, and the intrinsics (and distortion coefficients, if any) are known.
	\item The true distance between at least two of the vehicles in the group is known prior to commencement of flight.
	\item Communication delays between vehicles are ignored in the current scope of the work. Any information transmitted between vehicles is assumed to be received without delay or loss.
	\item During localization, we assume that the real world locations of the landmarks that constitute the map stay constant.
\end{enumerate}

\section{Vision based Collaborative Localization}

In this section, we detail the individual steps of the vision-based collaborative localization (VCL) algorithm. For reference, the general flow of the algorithm is depicted in figure \ref{fig:vcl}.

\subsection{Feature detection and matching}

The collaborative localization framework works on the basis of feature data: hence, the first step in the algorithm is to detect and describe salient features visible in the field of view of each camera. In our pipeline, we have utilized two kinds of detection/description methods: 

\begin{enumerate}
	\item CPU implementation: AKAZE - Accelerated KAZE features and M-LDB descriptors.
	\item GPU implementation: KORAL - Multi-scale FAST features and LATCH descriptors.
\end{enumerate}

AKAZE features are multi-scale features which are faster to compute compared to SIFT/SURF and demonstrate better accuracy than methods such as ORB \cite{alcantarilla2011fast}. The AKAZE algorithm creates a non-linear scale space pyramid to isolate salient features and then describes each keypoint through a 488-bit modified local binary (M-LDB) descriptor. Binary descriptors provide a significant advantage over their floating point counterparts (such as the ones used on SIFT/SURF) because of their low memory footprint, thus being applicable for algorithms of a cooperative nature that might require communication of feature data. In the second combination, we use a system known as KORAL. KORAL contains a modified version of the well-known FAST corner detection algorithm, which is robustified by applying it to every layer in a non-linear scale space pyramid. The features detected in this step are described through 512-bit binary vectors from the LATCH description algorithm \cite{levi2016latch}. As the next step to both these techniques, we currently utilize brute force matching on a Hamming distance metric in order to find matches between different sets of these binary descriptors. In general, the VCL framework is method-agnostic and can be adapted to any feature extraction or matching algorithm.

\subsection {Relative pose estimation}

The detection/matching step of the algorithm results in feature data: a combination of keypoint location coordinates in the image plane, and a binary descriptor data vectors corresponding to each location; and a list of common features between two or more views. The common feature data is then used to estimate relative poses between the vehicles as well as to create a sparse reconstruction of the environment. Furthermore, this relative pose estimation also comes into picture while performing inter-MAV estimation, as seen later.

Assuming two cameras are present, any projections of 3D points that are visible from both the images are related by the fundamental matrix. The analogue of fundamental matrix for calibrated cameras is the essential matrix ($\mathbf{E}$), a $3\times3$ matrix that encodes the epipolar geometry of the two views. $\mathbf{E}$ depends only on the rotation and translation between the two cameras, and can be estimated using point correspondences between the views from both cameras. In our approach, we utilize the 5-point algorithm presented in \cite{nister2003efficient} in order to estimate the essential matrix for a set of feature matches between two cameras.

\begin{eqnarray}
\mathbf{P}_R^{\top} \mathbf{E} \mathbf{P}_L = 0 \\
\mathbf{E} = [\mathbf{t}]_{\times}\mathbf{R}
\end{eqnarray}

We note here that the precision of the essential matrix depends heavily on the fidelity of the feature matches obtained as described in subsection A. When navigating in sparsely populated environments, at high speed, or when observing repetitive feature data (a common problem with textures in urban settings), feature matching is susceptible to a great amount of inaccuracy, resulting in false matches, which can then affect the relative pose estimation. A conventional way of solving this problem is by using the random sample consensus method (RANSAC): an iterative method that seeks to find outliers from the provided set of feature matches. Feature matches which are `farther' from the epipolar constraint as per the current consensus are considered to be inaccurate and discarded. RANSAC typically requires the choice of a parameter known as threshold ($T$), which determines the confidence. But as in our application, we would require the pose estimation to happen multiple times as the vehicles are in motion, the noise levels of the images/features might not be constant, and presetting the threshold parameter could result in degradation of performance over time.

\begin{figure}
	\centering
		\subfloat[Raw feature matches]{{\includegraphics[width = 3cm,height=5cm]{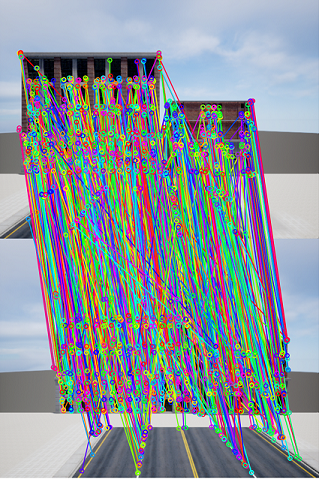}}}
		\hspace{1em}
		\subfloat[AC-RANSAC filtered matches]{{\includegraphics[width = 3cm,height=5cm]{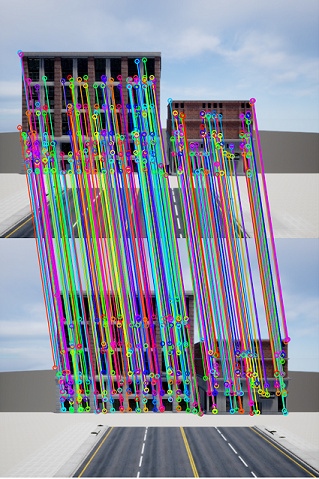}}}
	\caption{Demonstration of how AC-RANSAC helps with filtering out outliers in feature matches.}
	\label{fig:acransac}
\end{figure}

To avoid the problem of having to pick a sufficiently generalizable value for the threshold, we use a modified scheme known as the a-contrario RANSAC (AC-RANSAC). Proposed by Moisan and Stival \cite{moisan2004probabilistic} and demonstrated for computer vision applications in the subsequent work in \cite{moisan2012automatic}, a-contrario RANSAC is capable of choosing the value of parameter $T$ adaptively, based on the noise in the given data, thus enabling robust choice making in terms of the finding the right model, i.e., the essential matrix. Through this choice of value, the feature matches are filtered to remove outliers, and then the essential matrix is decomposed into rotation and translation. Figure \ref{fig:acransac} demonstrates how AC-RANSAC helps during essential matrix estimation by filtering out matches that don't adhere to a model.

\subsection {Map building}

Once a relative pose is known between two camera views, this information can be coupled with the feature matches to compute a sparse 3D reconstruction of the matched features. In the VCL algorithm, we use the DLT triangulation method \cite{hartley2003multiple} over all the feature matches to obtain a 3D map according the relative pose computed as described in the previous subsection. This map can then be made available to all the MAVs (for example, as a topic on ROS that individual vehicles can subscribe to) and is meant to be reused for feature tracking, 3D-2D correspondence computing as well as a source for performing scale factor recovery for subsequent reconstructions. We note here that for the first ever reconstruction and map building operation, according to assumption 2, having access to the distance between two MAVs helps remove the scale ambiguity problem with the reconstruction. The initial map, thus, is assumed to be metrically accurate: and this scale can be propagated through future updates. The globally available map data consists of two parts: one, a set of the 3D locations of all the points in the map, and two, the feature descriptors respective to all these points from the keyframes. Once an initial estimate of the 3D landmark positions is found with respect to the poses, both the poses and scene are jointly optimized through bundle adjustment for accuracy.

For a group containing more than two vehicles, we use an incremental reconstruction procedure. All MAVs capture images from their cameras, and a visibility graph of feature overlap is generated while isolating common features. The pair with the highest number of overlapping features is considered a seed pair and a first reconstruction is attempted. Once this reconstruction is generated, the feature observations of the other MAVs are incrementally included in this reconstruction. Finally, we use a fast bundle adjustment scheme to jointly optimize the poses and the scene landmarks.

\subsection{Intra-MAV localization}

Intra-MAV localization is the process that is performed by each MAV independently once a global map is distributed to all agents. Every time the MAV's onboard camera captures an image, the onboard algorithm attempts to track points from the 3D map that are still visible, identify their 2D projections in the image and use this information to estimate the vehicle's 6-DoF pose. This process involves applying the perspective-N-Point algorithm for a calibrated camera to estimate the relationship between a set of 3D points (tracked from the map) and their projections on the image plane. If $\mathbf{X_i}$ is a 3D point and $\mathbf{x_i}$ is its corresponding 2D projection, the following relationship encodes the pose of the camera responsible for this transformation.
\begin{equation}
\mathbf{x_i} = \mathbf{K}\begin{bmatrix}\mathbf{R} & \mathbf{t} \\ 0 & 1 \end{bmatrix} \mathbf{X_i}
\end{equation}

In our implementation, we combine a recently developed version of a perspective-3-point algorithm \cite{ke2017efficient} with another AC-RANSAC scheme, which is then applied to the tracked correspondences between the image and the 3D map, in order to estimate the position and orientation of the MAV. Once a pose estimate is computed, we attempt to refine the pose by minimizing the reprojection error as defined below:
\begin{equation}
\theta^* = \argmin_\theta \sum_{i}{\norm{\mathbf{x_i} - \pi(\mathbf{X_i}, \theta)}}
\end{equation}

$\pi$ encodes the camera projective transformation of a 3D point $\mathbf{X}_i$ onto the image plane for the pre-computed pose $\theta$, whereas $\mathbf{x}_i$ is the actual observation from the image at that time step. Through this step, we also obtain the solution quality encoded within a covariance matrix, which, combined with the final reprojection error, we then use to scale the measurement noise covariance for that MAV.

\begin{figure}
	\centering
	\subfloat[Example scenario where vehicle $V_1$ has better feature tracking from the map, whereas $V_2$ has minimal tracking, which would affect its localization adversely.]{\includegraphics[width = 7cm,height=5cm]{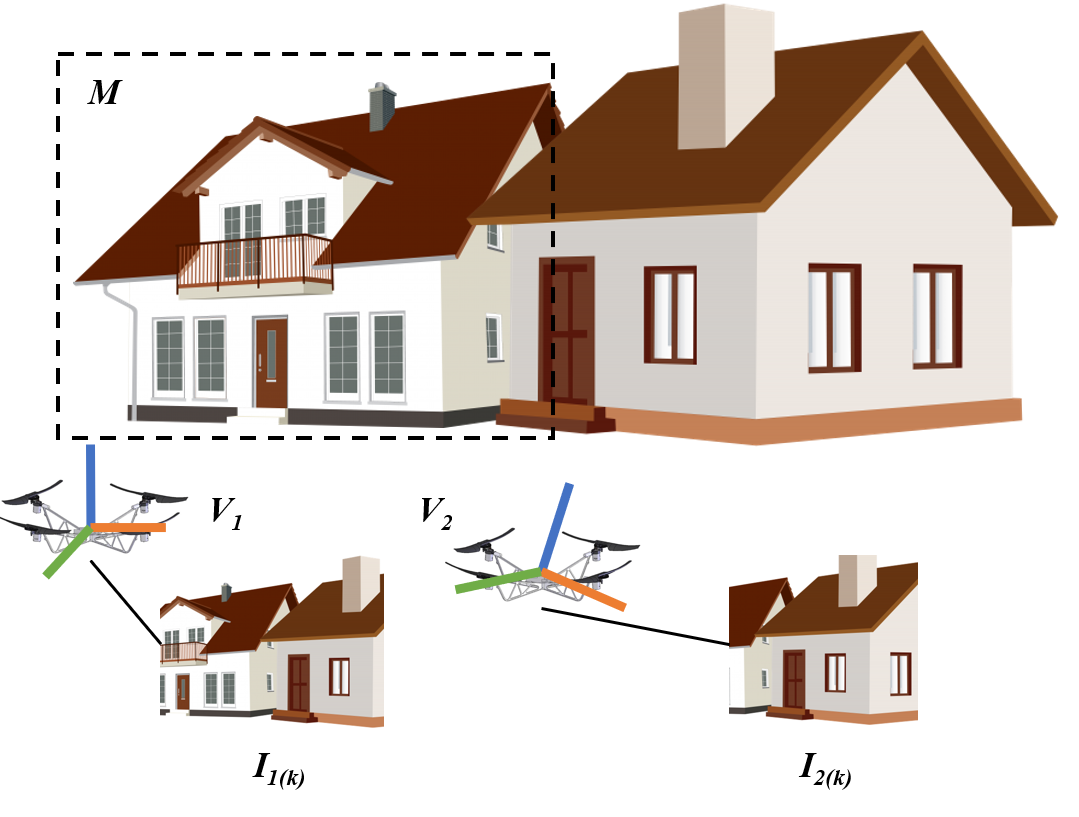}}
	\newline
	\subfloat[$V_2$ does not have sufficient overlap with map, whereas $V_1$ does]{{ \includegraphics[width = 3cm,height=4.5cm]{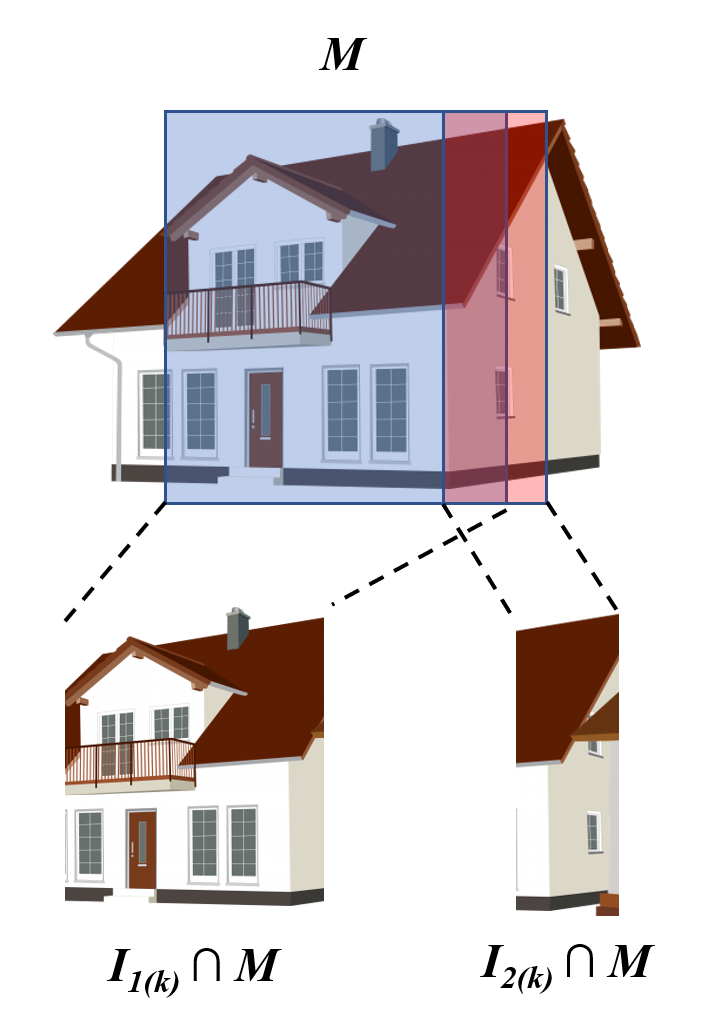} }} %
	\quad
	\subfloat[$V_2$ has significant overlap with $V_1$]{{ \includegraphics[width = 2.5cm,height=4.5cm]{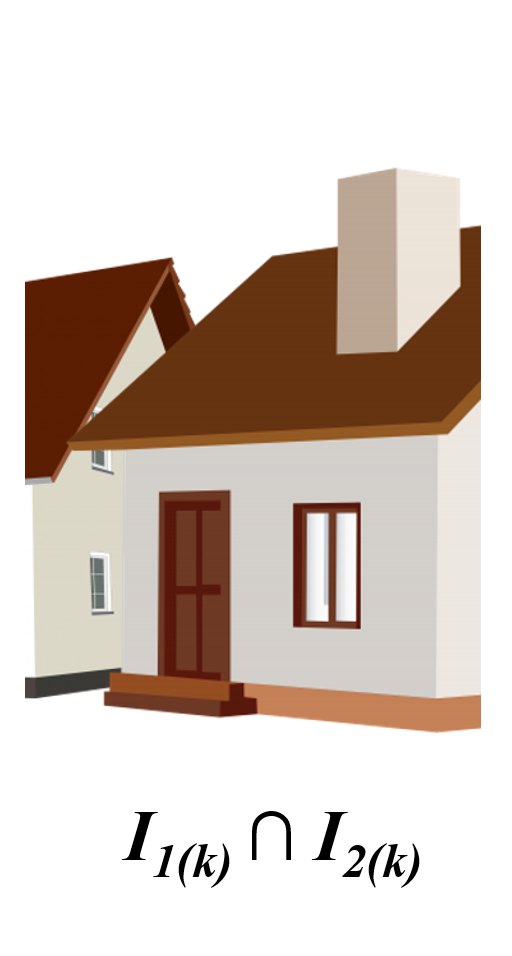} }} %
	\caption{Example scenario where intra-MAV localization may fail and inter-MAV localization can be beneficial.}
	\label{fig:intrabad}
\end{figure}

\subsection{Inter-MAV localization}

The success and the accuracy of intra-MAV localization depends entirely on the amount of overlap between the features in view of a certain vehicle and the features that comprise the global map. If an insufficient number of features are tracked by a vehicle, the error in pose estimation would increase drastically, which could then lead to drift. To assist with this, we capitalize on the existence of more than one vehicle, and attempt to use the collaborative nature to the advantage of localization. We call this step the 'inter-MAV' localization step. 

Figure \ref{fig:intrabad} shows a possible scenario where inter-MAV localization can be helpful. In this figure, for simplicity,we consider a case with only two MAVs: $V_1$ and $V_2$. At time instant $k$, vehicle $V_1$ has a good amount of overlap with the map $\mathbf{M}$ (blue region in Figure \ref{fig:intrabad}(b)), whereas vehicle $V_2$ does not (red region in Figure \ref{fig:intrabad}(b)). On the other hand, there is sufficient overlap between $V_1$ and $V_2$ independent of the map; with a non-zero overlap with the $\mathbf{M}$ (Figure \ref{fig:intrabad}(c)). Hence, the high confidence of $V_1$ in its own pose can be used to the advantage of $V_2$: we create a pipeline that enables $V_1$ to estimate the metric pose of $V_2$, which can then be fused with $V_2$'s own onboard estimate. 

The individual steps in this inter-MAV localization pipeline are described in detail below for an arbitrary pair of MAVs $V_i$ and $V_j$. The MAV requiring the inter-MAV estimate is called the `client' MAV and the one providing the estimate is called the `host' MAV.

\subsubsection{Relative pose estimation} 
As described in the first two subsections, first, the features visible from $V_i$ and $V_j$ are isolated and matched, and the essential matrix is used to compute a relative rotation and translation. This measurement of translation will only result in a unit vector with an arbitrary scale factor, thus is not sufficient for an accurate metric pose. This unit translation also results in a reconstruction of the features common to $V_i$ and $V_j$: a reconstruction that is scaled with an arbitrary, yet unknown scale factor. 

\subsubsection{Scale factor estimation}
The local map obtained through the common points between the images of $V_i$ and $V_j$ can be referred to as $\mathbf{M}'$. If an assumption is made that both $V_i$ and $V_j$ were able to localize using intra-MAV localization, albeit with varying degrees of accuracy, both $V_i$ and $V_j$ have a non-zero overlap with the existing map. Hence, it follows that there is a non-zero overlap of features between $\mathbf{M}'$ and $\mathbf{M}$.

In order to compute the right scale factor $\lambda$ for this reconstruction (one that matches the true scale of the global map), the frame of reference for the local map has to be considered. Within this local frame, the host MAV can be considered to be at the origin $[\mathbf{I}|\mathbf{0}]$ and the client MAV at $[\mathbf{R|t}]$, where $\mathbf{R}$ and $\mathbf{t}$ are the estimated relative rotation and translation between the host and client. If any two pairs of common features can be identified between the local and the global map, as their true 3D coordinates are already known from the global map, the ratio of the length of a line connecting the local coordinates to the length of one connecting the global coordinates is the true scale factor for the new map. Once this scale factor is known from the ratios, the relative pose is scaled to its right value, and the reprojection error is minimized to obtain a better estimate.

While comparing the inter-MAV local map and the global map, any wrong matches between the two sets of points can affect the estimation of the scale factor greatly. Although the goal of the AC-RANSAC scheme used during relative/individual pose estimation is typically to solve this very problem of removing outliers, in case of matching two maps during inter-MAV localization, the total number of points could be too low for a RANSAC scheme to act effectively. Hence, the VCL algorithm uses guided matching to ensure accuracy of matching between the two point sets.

\subsubsection{Guided Matching} 
The goal of this step is to find accurate matches between two maps: one being the typically denser global map, and the other, a sparse temporary map obtained by the matches between $V_i$ and $V_j$. Now, it can be recalled that the host MAV, which is responsible for generating relative poses has an acceptable degree of confidence in its own pose, which means that both the global map and local map are generated from confident poses. If the descriptors of the features that form the global and local maps can be assumed to represent two (virtual) images, the transformation between these two views is already known. Given this known transformation, it is possible to `guide' the matching towards inliers that adhere to the transformation \cite{hartley2003multiple}.

We can assume that the features belonging to the global map, or at least,a subset of them: are coming from an imaginary camera at pose $\mathbf{[I | 0}]$, and the features from the local map are from a camera that resides at the current pose of the host MAV, say, $[\mathbf{R' | t'}]$. Assuming finite overlap between these maps, there exists a relationship between these two cameras, and therefore, the maps, which can be described using the fundamental matrix as
\begin{equation}
\mathbf{F} = \mathbf{K_2^{-\top} R' K_1 (K_1 R'^{\top} t')}_{\times}
\end{equation}

Given that $\mathbf{R'}$ and $\mathbf{t'}$ are known, it is trivial to compute the fundamental matrix according to equation 7. Once $\mathbf{F}$ is known, any matches that are not adherent to the proper epipolar geometry (as represented in equation 8) can be considered outliers and discarded. 
\begin{equation}
\mathbf{x}_2^{\top} \mathbf{F} \mathbf{x}_1 = 0
\end{equation}

\begin{figure}
	\centering
	\includegraphics[width =7cm,height=8cm]{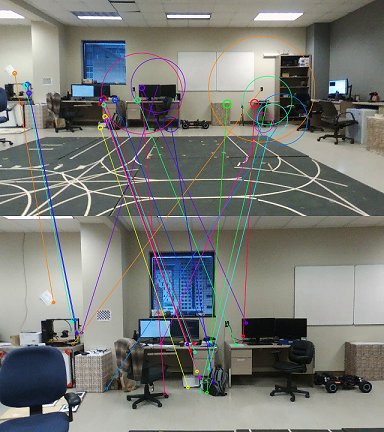}
	\caption{Guided matching to remove outliers: false matches are seen to have a high epipolar error, corresponding circles seen to have a much larger radius than those of the inlier matches.}
	\label{fig:guidedmatching}
\end{figure}

Figure \ref{fig:guidedmatching} shows an example of guided matching and how it helps determine outliers. The matches computed by the algorithm between two maps are drawn as colored lines, with the radius of the circles at the endpoints indicating the corresponding epipolar error. In practice, no match corresponds to exactly zero epipolar error, so a pixel based threshold on the error helps discard all matches that exhibit a high error. The inliers are the matches with endpoint circles of small radii.

Intra-MAV estimation usually suffers in accuracy when there are not enough features to be tracked from the original map. In such cases, inter-MAV estimation can be helpful as it utilizes common features between the MAVs at that instant and does not require multiple observations over time. Scale recovery in the inter-MAV estimation step requires a minimum of only two pairs of accurate matches between the local and global map, as opposed to intra-MAV estimation, which requires a significantly higher number of tracked features for better accuracy.

\subsection{Uncertainty estimation}

One of the critical parts of localization is to estimate not only the position and orientation of a vehicle, but also estimate the uncertainty of the estimated pose, which is usually described through a covariance matrix. This covariance is conventionally propagated through a Kalman filter framework while predicting and updating poses. In our algorithm, we also follow a Kalman Filter predict-update procedure which will be described later, but in order to accurately propagate the measurements and associated uncertainty through the filter, initially, it is important to describe the accuracy of each measurement received.

In both the inter and intra-MAV estimation steps, a final refinement step is performed through a non-linear least squares method where the algorithm attempts to correct the pose by minimizing the reprojection error. For a projective transformation $\pi$, the cost function that is being optimized, i.e, sum of all residuals: can be written for $m$ features as

\begin{gather}
f(\theta) = \frac{1}{2} \norm{\mathbf{r}(\theta)}^2  \\
\mathbf{r} = \begin{bmatrix} r_1 & r_2 & ... & r_m \end{bmatrix} \textrm{where} r_i(\theta) = x_i - \mathbf{\pi}(X_i, \theta) \nonumber
\end{gather}

The Jacobian and the Hessian of this function can be evaluated to be:

\begin{align}
\nabla f(\theta) & = \sum_{i} r_i(\theta) \nabla r_i(\theta) \nonumber \\
& = \mathbf{J}(\theta)^T \mathbf{r}(\theta) \\ \nonumber 
\end{align}
\begin{align}
\nabla^2 f(\theta) & = \sum_{i} \nabla r_i(\theta) \nabla r_i(\theta)^T + \sum_{i} \nabla^2 r_i(\theta) \\
& \approx \mathbf{J}(\theta)^T \mathbf{J}(\theta)
\end{align} 

When a solution is close to a local minimum, the effect of the second order terms in equation 11 is minimized, and hence they can be ignored in terms of their contribution towards the objective. Once the second order terms are ignored, it can be seen that the outer product of this Jacobian matrix at the final optimum with itself is an approximation of the Hessian matrix of the solution (eq. 12). For a non-linear multidimensional least squares error near the optimum, the inverse of this Hessian matrix is an approximation of the covariance matrix of the reprojection errors \cite{press1996numerical}. Hence, the approximate covariance of the solution can be expressed as

\begin{equation}
\mathbf{\Sigma} = (\mathbf{J}^{\top}\mathbf{J})^{-1}
\end{equation}

We note here that $\mathbf{\Sigma}$ in equation 13 does not necessarily translate into an uncertainty in the real position/orientation values directly: it merely expresses the quality of the solution and the possible uncertainty around the local surface at the point of convergence. This value is still a function of the reprojection errors and not of the rotation/translation parameters: in case the solution is a local minimum, the estimated covariance could still be low although the pose estimate is not very true to the actual value. So in order to express the pose uncertainty more accurately, we scale this covariance artificially with the reprojection error obtained for that pose estimate.

\begin{equation}
\mathbf{R} = (\mathbf{J}^{\top}\mathbf{J})^{-1} * \epsilon_{r} 
\end{equation}

\subsection{Kalman Filter and Outlier rejection}

As a final step, the raw measurements obtained during the course of the VCL framework are propagated through a Kalman filter framework. The VCL scheme being a purely vision based localization system without augmentation from other sensors like IMUs, the cameras are essentially considered to be replacements for the vehicles in terms of poses. Due to this reason, the primary function of the Kalman filter in the VCL framework is for smoothing and outlier rejection, utilizing a constant velocity model as the process. This filtering scheme can be modified to incorporate a more complex vehicle model, or include IMU measurements as an extension to this work. 

Each MAV is responsible for running its own internal recursive estimation scheme through a Kalman filter which maintains a running estimate of the mean and covariance of its state. At every instant an image is received, it is expected that the MAV computes an intra-MAV pose estimate for itself. Once computed, the obtained measurement is used to correct the state and covariance of that particular MAV. If a measurement obtained at time step $k$ is denoted as $\mathbf{z}_{k}^i$,
\begin{equation}
\mathbf{z}_{k}^i = \mathbf{h}(\mathbf{x}_{k}^i) + \mathbf{n_{k}^i}
\end{equation}
The measurement is then used to correct the predicted pose at time step $k$, where the measurement noise covariance is inflated as discussed in equation (14).
\begin{align}
\mathbf{P}_{k|k-1}^i = \mathbf{A}_k^i \mathbf{P}_{k-1|k-1}^i \mathbf{A}_k^{i^\top} + \mathbf{Q}_{k}^i \\ \nonumber
\mathbf{S}_{k}^i = \mathbf{H}_{k}^i \mathbf{P}_{k|k-1}^i \mathbf{H}_{k}^{i^\top} + \mathbf{R}_{k}^i \\ \nonumber
\mathbf{P}_{k|k}^i = (\mathbf{I} - \mathbf{KH})\mathbf{P}_{k|k-1} 
\end{align}  
Before using the obtained pose value in the measurement update, it is beneficial to determine the likelihood of the measurement being an outlier. One standard way of performing this check is through what is known as a Chi-squared gating test. The VCL algorithm uses this test to detect and reject very noisy observations. Every time a new measurement for the MAV pose is available, the Mahalanobis distance is computed between the expected measurement and the actual measurement as 
\begin{equation}
\gamma_k = (\mathbf{z}^i_{k} - \hat{\mathbf{z}}^i_{k})^{\top} \mathbf{S}^{-1} (\mathbf{z}^i_{k} - \hat{\mathbf{z}}^i_{k})
\end{equation}
The rank of $\mathbf{S}$, which is computed as part of the Kalman filter update sequence in (16), determines the system's degrees of freedom, which for a full 6 DoF pose, would be 6. Assuming the process/measurement noises are Gaussian distributed (which is usually the case), $\gamma_k$ should be Chi-square distributed with 6 degrees of freedom. It is possible to decide upon a certain probabilistic threshold value: which, when exceeded, makes a candidate measurement an outlier. Hence, if $\gamma_k$ exceeds the $\alpha$-quantile of the Chi-squared distribution, the measurement can be treated as an outlier and discarded \cite{chang2014robust}; \cite{mirzaei2008kalman}).

In the context of inter-MAV estimation, the responsibility of computing both the state and the covariance of a client MAV is taken up by the host MAV. Once a relative measurement between the host and client, denoted as $\mathbf{z}_{k}^{i,j}$ is available, host MAV $V_i$ uses this relative measurement in conjunction with its own state estimate to compute the state of MAV $V_j$ as follows.
\begin{equation}
\mathbf{x}_k^{j'} = \mathbf{x}_{k}^{i} + \mathbf{M}_{k}^{i,j} \mathbf{z}_{k}^{i,j} \\
\end{equation}

In the VCL scheme, the measurement computed by the host is the pose of the client directly, hence $M$ in equation (18) is an identity matrix. It is important to remember that when $V_i$ attempts to compute the uncertainty of $V_j$, this is in combination with $V_i$'s own uncertainty. Hence, the covariance matrix that $V_i$ has estimated for itself should be propagated into any other relative measurements attempted by $V_i$. When $V_i$ computes a relative measurement to $V_j$ at time instant $k$ with measurement noise covariance $\mathbf{R}^{ij}_k$, the corresponding state covariance for $V_j$: $\mathbf{P}^{j'}_k$ can be calculated as:
\begin{equation}
\mathbf{P}^{j'}_k = \mathbf{H}^{j'}_k \mathbf{P}^i_{k|k} \mathbf{H}^{j'^{\top}}_k + \mathbf{R}^{ij}_k
\end{equation}

\subsection{Data fusion}

\begin{figure}
	\includegraphics[width = 8cm,height=5cm]{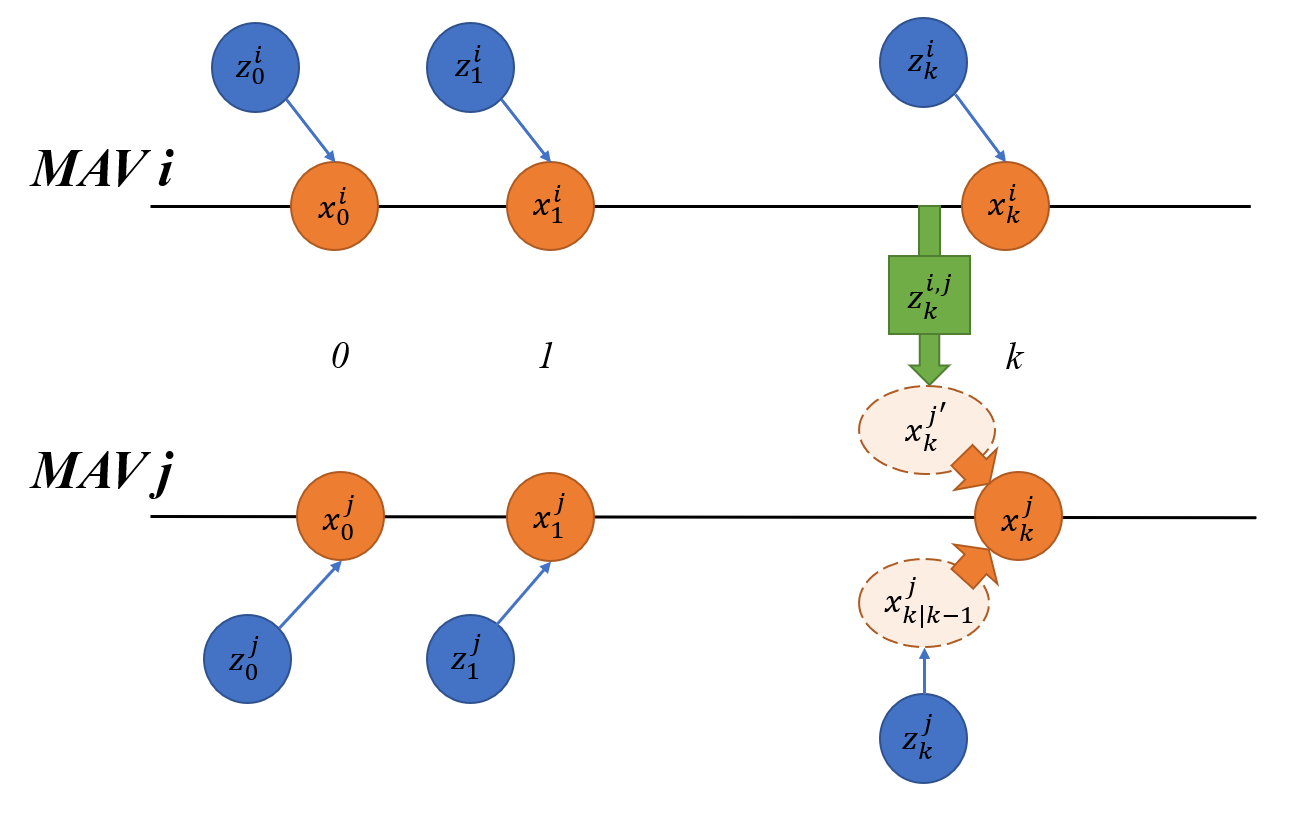}
	\caption{A pictorial representation of data fusion between inter-MAV and intra-MAV localization. At time step $k$, MAV $i$ attempts to correct the pose of MAV $j$ by generating a relative measurement, which is then fused by MAV $j$ with its own onboard estimate.}
	\label{fig:fusion}
\end{figure}

One of the strengths of a collaborative localization routine is the ability to fuse estimated data from multiple sources, thus attempting to compute more robust pose estimates. In the case of the VCL algorithm, we can intuitively see one vehicle has its own `intra-MAV' estimate, and possibly multiple `inter-MAV' estimates computed by some other vehicles in the group. For simplicity, if we consider two vehicles in the group, the pose computed by $V_i$ for $V_j$ needs to be fused with the onboard estimate of MAV $V_j$ itself (an example scenario is depicted in Figure \ref{fig:fusion}). A conventional way of data fusion is to include both sources in a Kalman filter update framework, but here, it should be noted that this relative measurement and the measurement that resulted in the local pose of MAV $j$ have common sources of information, i.e., the map data and the features. These two MAVs could also have communicated pose data in the past, which makes these estimates correlated. But because these cross-correlation parameters are not kept track of, the correlations are treated as unknown: which makes the conventional EKF update step result in inconsistent and erroneous estimates. To handle this problem, we utilize the approach described in \cite{carrillo2013decentralized}: which attempts to fuse multiple estimates using covariance intersection. 

Covariance intersection (CI), first presented in the seminal paper \cite{julier1997non}, is an elegant solution for the fusion of estimates with unknown correlations. The CI algorithm expresses the covariance of the fused estimate as a combination of the individually estimated covariances. Depending on the confidence in each estimate or a desired final statistic, each individual covariance can be weighted by a scalar value. In the case of the VCL framework, each estimate is already described by its own confidence coming either from onboard the same vehicle requiring fusion, or the host vehicle that generated a pose for the client. At time instant $k$, assume that the individual estimate of $V_j$ is a state-covariance pair with state $\hat{\mathbf{x}}_k^j$ and covariance $\hat{\mathbf{P}}_k^j$. For the same time instant, $V_i$ computes another state-covariance pair for the pose of $V_j$, represented as $\hat{\mathbf{x}}_k^{ij}$ and $\hat{\mathbf{P}}_k^{ij}$. Then the CI algorithm can be used to compute a state and covariance pair of a fused estimate as below.
\begin{gather}
\mathbf{P}_{k}^j = {\Big[\omega(\mathbf{P}_{k}^j)^{-1} + (1-\omega)(\mathbf{P}_{k}^{ij})^{-1}\Big]}^{-1} \\
\mathbf{x}_{k}^j = \mathbf{P}_{k}^j \Big[{\omega}(\mathbf{P}_{k}^j)^{-1} \mathbf{x}_{k}^j + (1-{\omega})(\mathbf{P}_{k}^{ij})^{-1} \mathbf{x}_k^{ij} \Big] 
\end{gather}

Where $\omega$ is a parameter that is computed such that the trace of the combination of the covariances being fused is minimized: this can be expressed as in equation 22. 
\begin{equation}
\argmin_\omega Tr{\Big[{\omega}(\mathbf{P}_{k}^j)^{-1} + (1-{\omega})(\mathbf{P}_{k}^{ij})^{-1}\Big]}^{-1}
\end{equation}

Another elegant property of the CI algorithm is that, while being derived from a geometric viewpoint, it can also be expressed as a matrix- and scalar-weighted optimization problem. This property allows the CI algorithm to be extended to the fusion of higher dimensional state vectors and thus, an arbitrary number of estimates. Consequently, in the VCL framework, fusion can be performed between more than two sources of data, where for $k$ sources, the covariance matrices are weighted by an array of $k$ weighting factors $\omega_1, \omega_2, ..., \omega_k$ such that
\begin{equation}
\omega_1 + \omega_2 + ... + \omega_k = 1
\end{equation}

\begin{figure}
	\begin{center}
		\centerline{\includegraphics[width = 8cm]{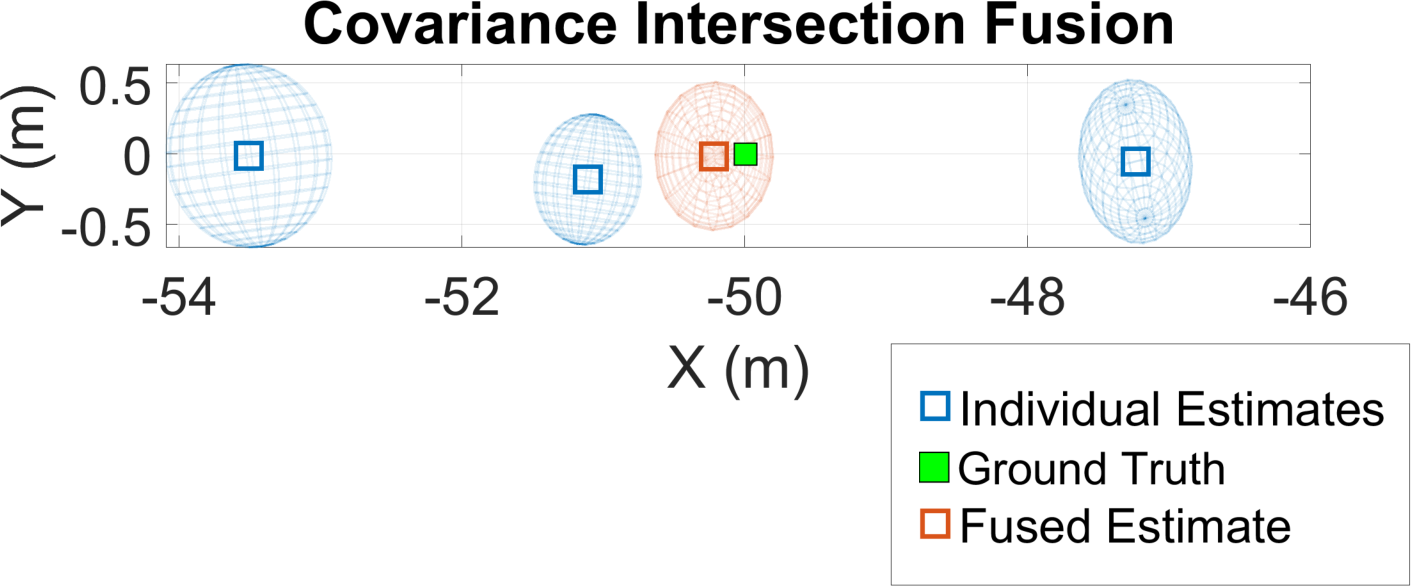}}
		\caption{Example of covariance intersection fusion of data from three sources.}
		\label{fig:covintexample}
	\end{center}
\end{figure}

For $k$ sources of data, equation 22 can be extended into a multi-variable minimization problem of finding a set of weights that minimize the weighted sum of the traces of all covariance matrices involved. Figure \ref{fig:covintexample} is a visual depiction of covariance intersection fusion for data from 3 sources, from an experiment where 3 vehicles were responsible for estimating the position of a fourth on the X axis. At that particular time step, none of the estimates are particularly close to the ground truth, but the level of confidence exhibited by the closest estimate is higher compared to the others; which results in the covariance intersection algorithm computing the right combination of weights to result in a fairly accurate fused estimate. 

\subsection{Map updates}

While the formulation and focus of the VCL algorithm is mainly on localization, mapping is an essential part of the process. In certain situations that could be part of the application, all the MAVs may have to move away from an initial map, which would necessitate an update of the global map in order to maintain localization. While inter-MAV localization is able to assist with the case of specific MAVs leaving the area of the mapped scene, all the vehicles navigating to a different area would require a new map. Hence, the same principles that allow for inter-MAV localization, i.e, relative pose estimation and scale recovery, can be used for building a new global map. This process can be invoked when the tracked feature count falls under a certain threshold for all the vehicles in the group or any other condition that is deemed appropriate. The steps involved in a map update follow the steps in the inter-MAV localization closely: 

\begin{algorithm}
	\caption{VCL algorithms: sample for two MAVs}
	\begin{algorithmic}
		\Procedure{buildMap}{$x_i$, $I_1$, $I_2$}
		\State $i_1, i_2$ $\gets$ detectFeatures$(I_1, I_2)$
		\State $\bar{i_1}, \bar{i_2}$ $\gets$ matchFeatures$(i_1, i_2)$
		\State $\mathbf{E}$ $\gets$ ACRANSAC($\bar{i_1}, \bar{i_2}, \mathbf{K_1}, \mathbf{K_2}$)
		\State $\mathbf{R}, \mathbf{t}$ $\gets$ SVD$(\mathbf{E})$
		\State $\mathbf{M}$ $\gets$ reconstruct$(\bar{i_1}, \bar{i_2}, [I|0], [\mathbf{R}, \mathbf{t}])$ \Comment{$M$: Global map}
		\EndProcedure%
		\\
		\Procedure{localizeIntraMAV}{$I_k$, $\mathbf{K_k}$, $\mathbf{M}$}
		\State $i_k$ $\gets$ detectFeatures$(I_k)$
		\State $\bar{i_1} \gets$ trackFeatures$(i_k, \mathbf{M})$
		\State $\mathbf{R}, \mathbf{t}$ $\gets$ PNP($\bar{p_k}, \mathbf{M}, \mathbf{K_1})$
		\State $\mathbf{z_k}, \mathbf{R_k}$ $\gets$ refinePose$(\mathbf{R}, \mathbf{t}, \mathbf{M})$
		\State $\mathbf{x}_{k}, \mathbf{P}_{k} \gets$ updateState$(\mathbf{z}_k, \mathbf{R}_k)$
		\State \textbf{return} $\mathbf{x}^{j}_k, \mathbf{P}^{j}_k$
		\EndProcedure%
		\\
		\Procedure{localizeInterMAV}{$\mathbf{x}_i$, $I_i$, $I_j$}
		\State $i_i, i_j$ $\gets$ detectFeatures$(I_i, I_j)$
		\State $\bar{i_i}, \bar{i_j}$ $\gets$ matchFeatures$(i_i, i_j)$
		\State $\mathbf{E}$ $\gets$ ACRANSAC($\bar{i_1}, \bar{i_2}, \mathbf{K_i}, \mathbf{K_j}$)
		\State $\mathbf{R}, \mathbf{t}$ $\gets$ SVD$(\mathbf{E})$
		\State $\mathbf{M'}$ $\gets$ reconstruct$(\bar{i_i}, \bar{i_j}, [I|0], [\mathbf{R}, \mathbf{t}]$) \Comment{$\mathbf{M}'$: Local reconstruction}
		\State $\mathbf{m}_{map} \gets$ matchFeatures$(\mathbf{M}', \mathbf{M})$ \Comment{$\mathbf{M}$:= Global map}
		\State $\lambda \gets$ recoverScale$(\mathbf{M}', \mathbf{M}, \mathbf{m}_{map})$
		\State $\mathbf{t} \gets \mathbf{t} * \lambda$
		\State $\mathbf{z}^{i,j}_k = [\mathbf{R}, \mathbf{x_i} + \mathbf{t}]$
		\State $\mathbf{z}^{i,j}_k, \mathbf{R}^{i,j}_k$ $\gets$ refinePose$(\mathbf{R}, \mathbf{t}, \mathbf{M}')$
		\State $\mathbf{x}^{j'}_k, \mathbf{P}^{j'}_k \gets eqn(18), (19)$
		\State \textbf{return} $\mathbf{x}^{j'}_k, \mathbf{P}^{j'}_k$
		\EndProcedure%
		\\
		\Procedure{fuseInterIntra}{$(\mathbf{x}^j_k, \mathbf{P}^j_k), (\mathbf{x}^{j'}_k, \mathbf{P}^{j'}_k)$}
		\State $\mathbf{P_A} \gets \mathbf{P}^j_k$
		\State $\mathbf{P_B} \gets \mathbf{P}^{j'}_k$
		\State $\omega \gets \argmin_\omega Tr(\omega \mathbf{P_A}^{-1} + (1-\omega)\mathbf{P_B}^{-1})$
		\State $\mathbf{P}^{j*}_k \gets (\omega \mathbf{P_A}^{-1} + (1-\omega)\mathbf{P_B}^{-1})^{-1}$
		\State $\mathbf{x}^{j*}_k \gets \mathbf{P}_{k}^{j*} (\omega \mathbf{P_A}^{-1} \mathbf{x}_{k}^j + (1-{\omega})\mathbf{P_B}^{-1} \mathbf{x}_k^{j'})$
		\EndProcedure
	\end{algorithmic}
\end{algorithm}

\begin{enumerate}
\item Perform feature matching and relative pose estimation to result in a new, scale-ambiguous reconstruction.
\item Perform matching between the new reconstruction and the existing global map in order to scale the new reconstruction accurately.
\item Perform fast bundle adjustment to jointly optimize poses and map points and then, either replace the global map with the new one, or append new points to existing map.
\end{enumerate}

\begin{figure}
	\centering
	\captionsetup[subfigure]{justification=centering}
	\subfloat[Simulated city environment in Unreal Engine/AirSim]{\includegraphics[width = 4cm,height=3cm]{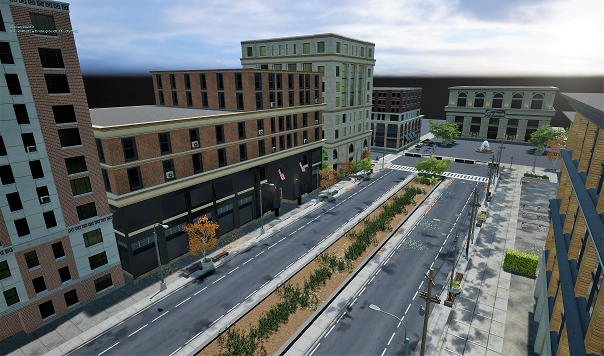}}
	\quad
	\subfloat[Parrot Bebop 2 quadrotors used in real flight experiments]{\includegraphics[width = 4cm,height=2cm]{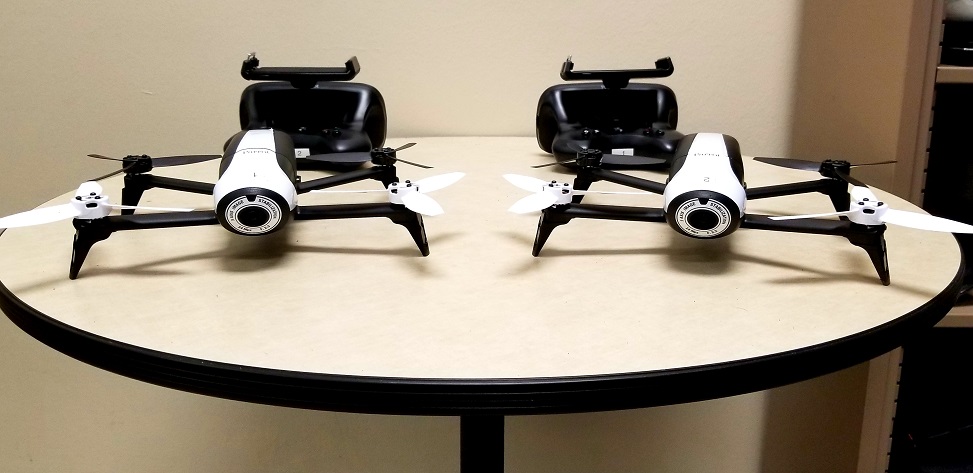}}
	\caption{Implementation details for simulation and real experiments}
	\label{fig:impl}
\end{figure}

\section {Implementation}

The collaborative localization framework has been written in C++, where we utilize various open-source libraries available in OpenCV \cite{opencv_library} and OpenMVG \cite{openMVG} in order to implement feature detection, matching, AC-RANSAC and PNP pose estimation. Ceres libraries \cite{ceres-solver} were utilized to refine reconstructions and estimated poses, as well as for estimating covariances of the solutions, and dlib \cite{dlib09} was used to perform optimization for the covariance intersection. The algorithm was run on an Intel NUC computer with an Intel i7-6770HQ processor, 32 GB of RAM and an NVIDIA GTX 1080 GPU used as an external GPU.

The collaborative localization algorithm has been tested on datasets obtained from both simulated and real flight tests. For the simulations, we use Microsoft AirSim as our platform. AirSim is a recently developed UAV simulator built as a plugin for Unreal Engine, a AAA videogame engine with the capability to render high resolution textures, realistic lighting, soft shadows, extensive post-processing etc. in order to bring simulated environments closer to real life fidelity. As our technique is heavily dependent on computer vision, using AirSim enabled testing in close-to-real-life situations. Each MAV simulated within AirSim had a forward facing monocular camera, and onboard images were captured at approximately 5 Hz with a resolution of 640x480. We created an urban environment in Unreal Engine within which the MAVs were flown through different trajectories (example picture from the environment can be seen in Figure \ref{fig:impl}(a)). The images from the onboard cameras and ground truth were recorded and the collaborative localization algorithm was tested offline on this pre-recorded data. For the real life tests, we used two Parrot Bebop 2 quadrotors (Figure \ref{fig:impl}(b)). Videos from the forward facing monocular cameras were recorded onboard the vehicles at a 1280x720 resolution, and then processed offline. Due to limitations of the Bebop 2 platform, it was not possible to capture both GPS/IMU updates along with high-frequency images in a timestamped way, hence, we limit our quantitative error analysis to simulation. The assumption regarding initial distance between the drones  was satisfied by starting the flights from pre-marked positions. A quick summary of our implementation and results can be found in a supplementary video\footnote{Supplementary video can be found at\\\url{https://www.youtube.com/watch?v=LvaTOWuTOPo}} .

\section{Results and Discussion}

\subsection{Intra-MAV localization: simulation}

\begin{table}
	\centering
	\begin{tabular}{c | c | c | c | c}
		\hline
		MAV ID & Error & X (cm)  & Y (cm)  & Z (cm)  \\
		\hline
		\multirow{2}{*}{1}& RMSE & 2.63 & 4.75 & 4.94 \\ 
		&	Max & 66.45 & 52.45 & 98.23   \\
		\hline
		\multirow{2}{*}{2}& RMSE & 2.77 & 4.88 & 4.70 \\
		& Max & 62.41 & 65.43 & 71.28   \\
		\hline
		\multirow{2}{*}{3}& RMSE & 2.39 & 3.85 & 4.49 \\
		& Max & 44.99 & 65.29 & 84.11   \\
		\hline	
		\hline
	\end{tabular}
	\caption[Table caption text]{RMS/maximum absolute errors for position estimates of three MAVs in AirSim}
\end{table}

\begin{figure}
	\centering
	\captionsetup{justification=centering}
	\includegraphics[width = 6cm,height=6cm]{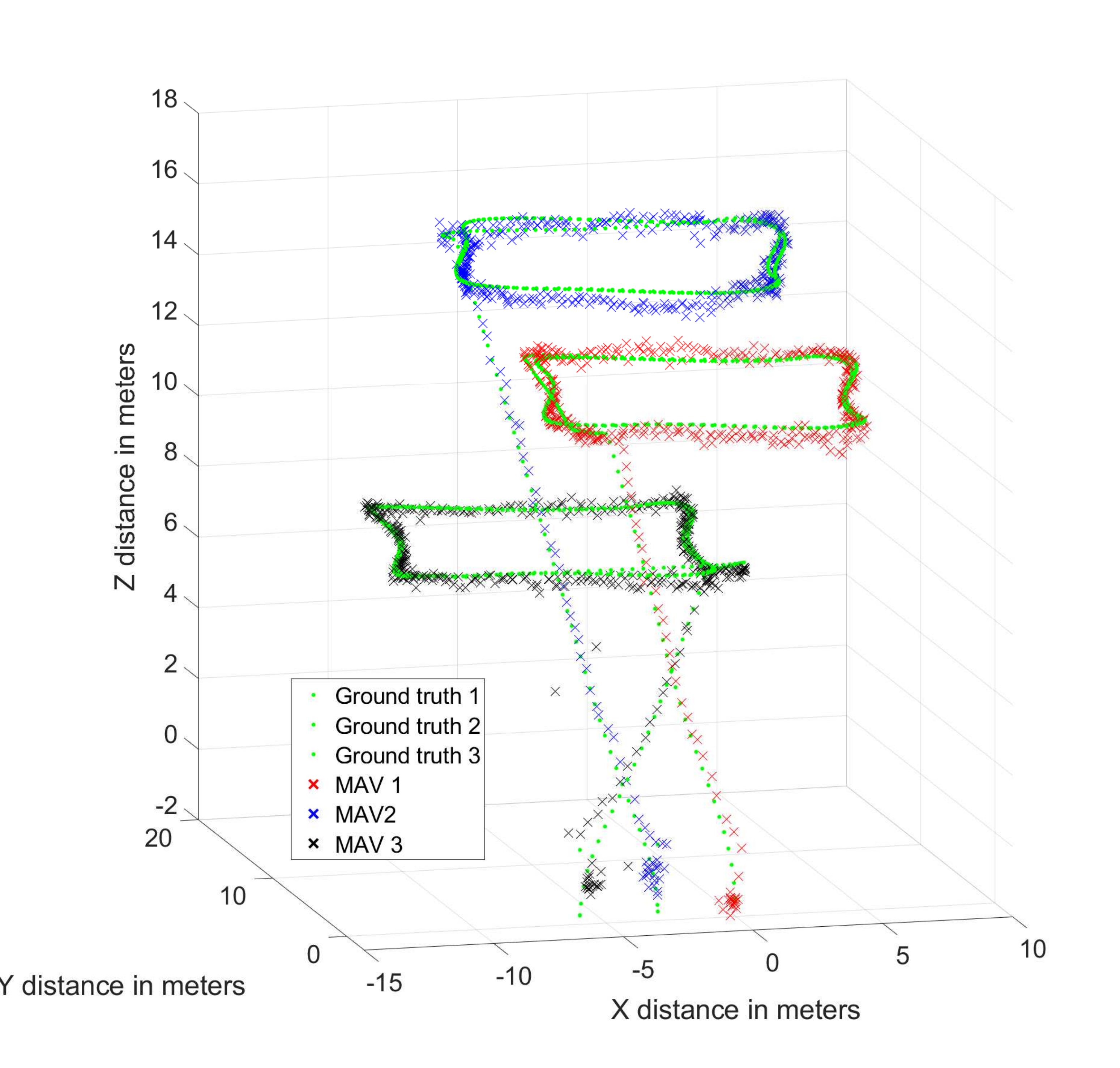}
	\caption{VCL position estimates for three MAVs navigating within AirSim - intra-MAV localization only.}
	\label{fig:3square}
\end{figure}

As a first step, we initialized three MAVs in the simulation environment, which were then commanded to take off and fly in square-like trajectories at different altitudes, while onboard camera images were recorded. Within this dataset, the algorithm was asked to build a global map using images from the three vehicles and only intra-MAV localization was tested for each vehicle. Figure \ref{fig:3square} shows the three estimated trajectories of the vehicles along with the ground truth. Table 1 shows the RMS errors of the VCL estimates for the three vehicles compared to the respective ground truth positions.

\begin{figure}
	\centering \subfloat[X-Y positions of two Bebop 2 quadrotors moved through a trajectory of known, marked dimensions. Ground truth plotted in black.]{{\includegraphics[width = 4cm,height=3cm]{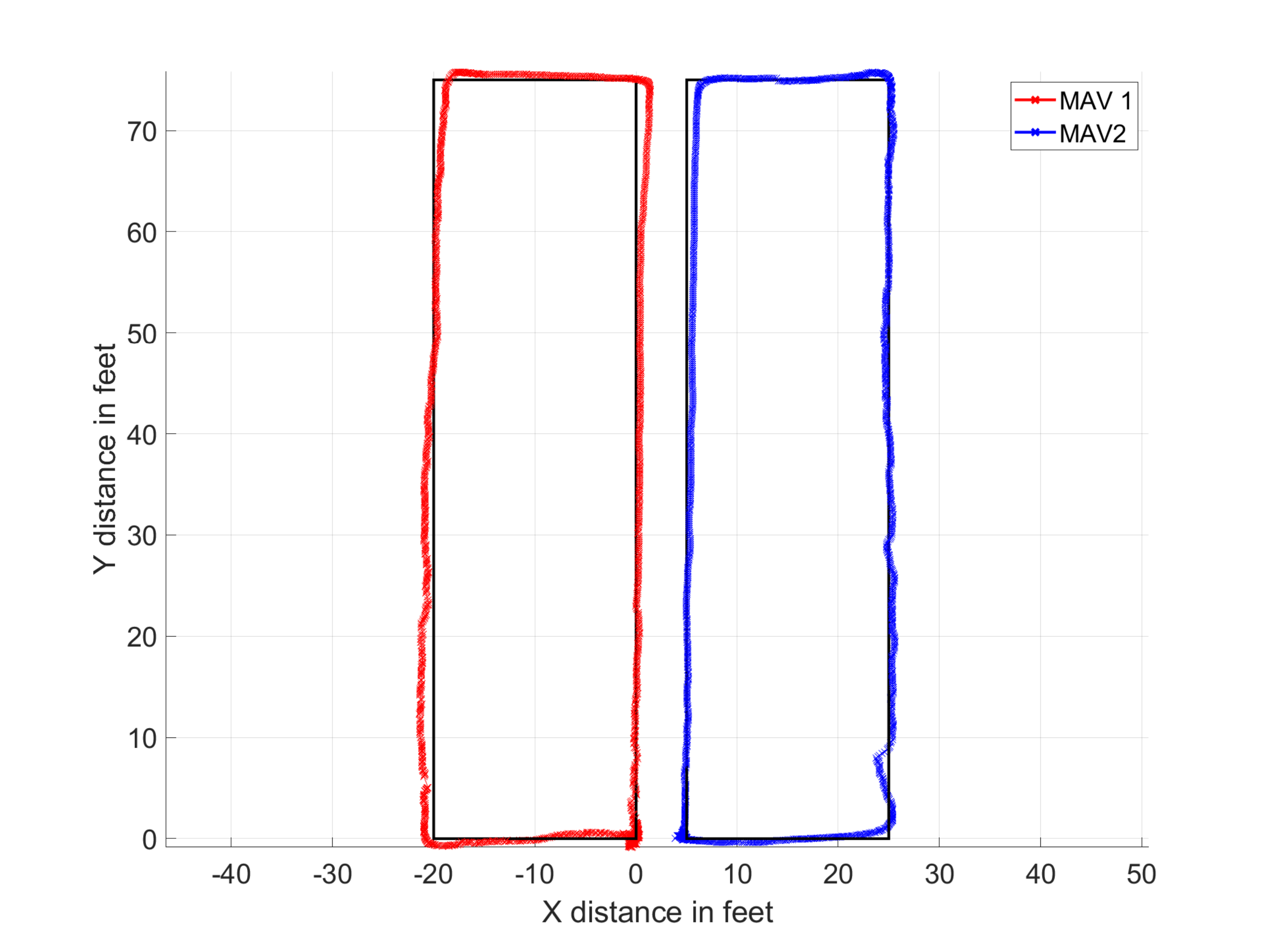}}}
	\hspace{0.1em}
	\centering \subfloat[X-Y positions of two Bebop 2 quadrotors flown through rectangular trajectories. GPS position estimates plotted in dotted black.]{{ \includegraphics[width = 4cm,height=3cm]{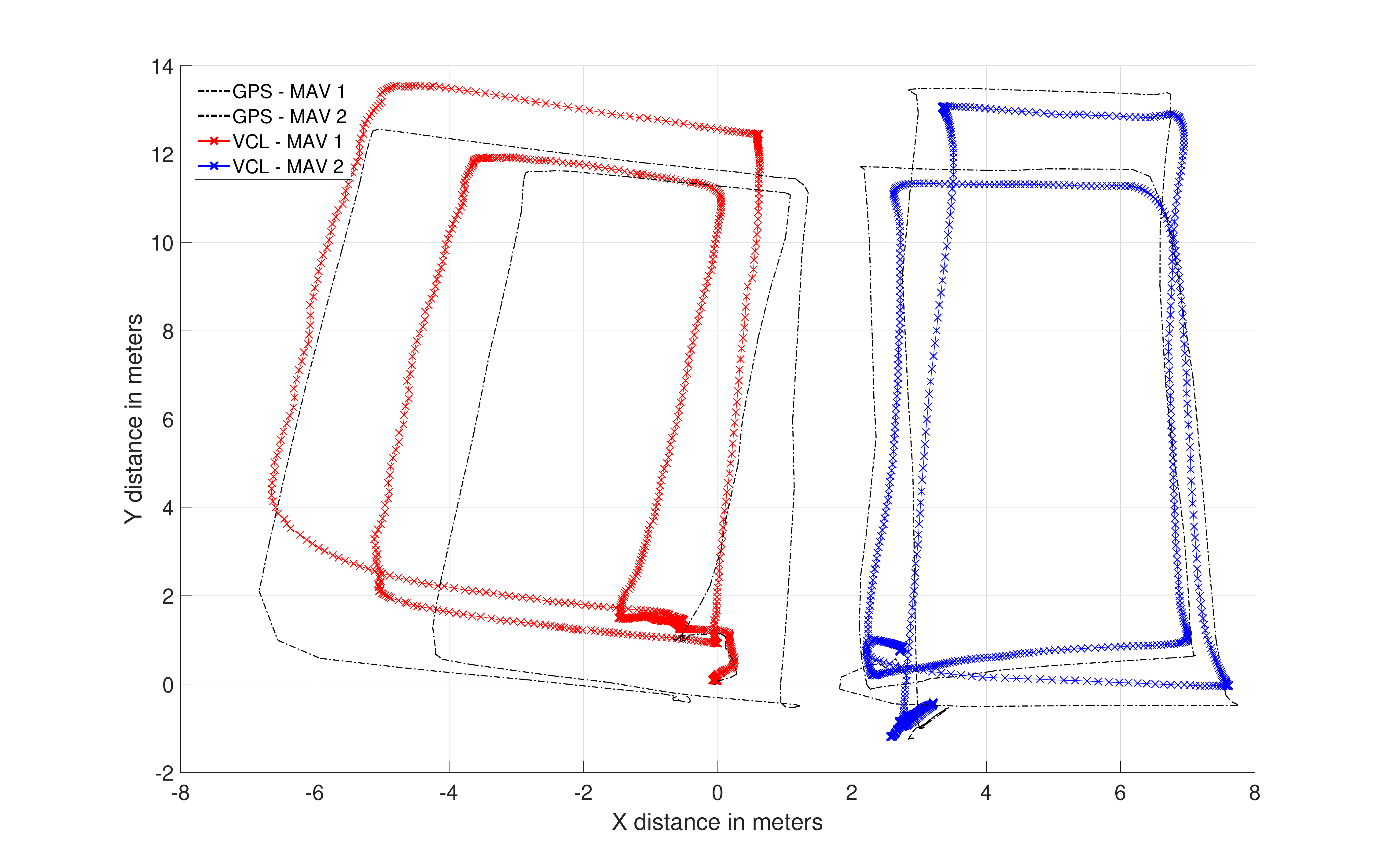} }} %
	\caption{Intra-MAV localization in real experiments}
	\label{fig:intrareal}
\end{figure} 

\begin{figure}
	\centering
	\includegraphics[width = 9cm,height=4cm]{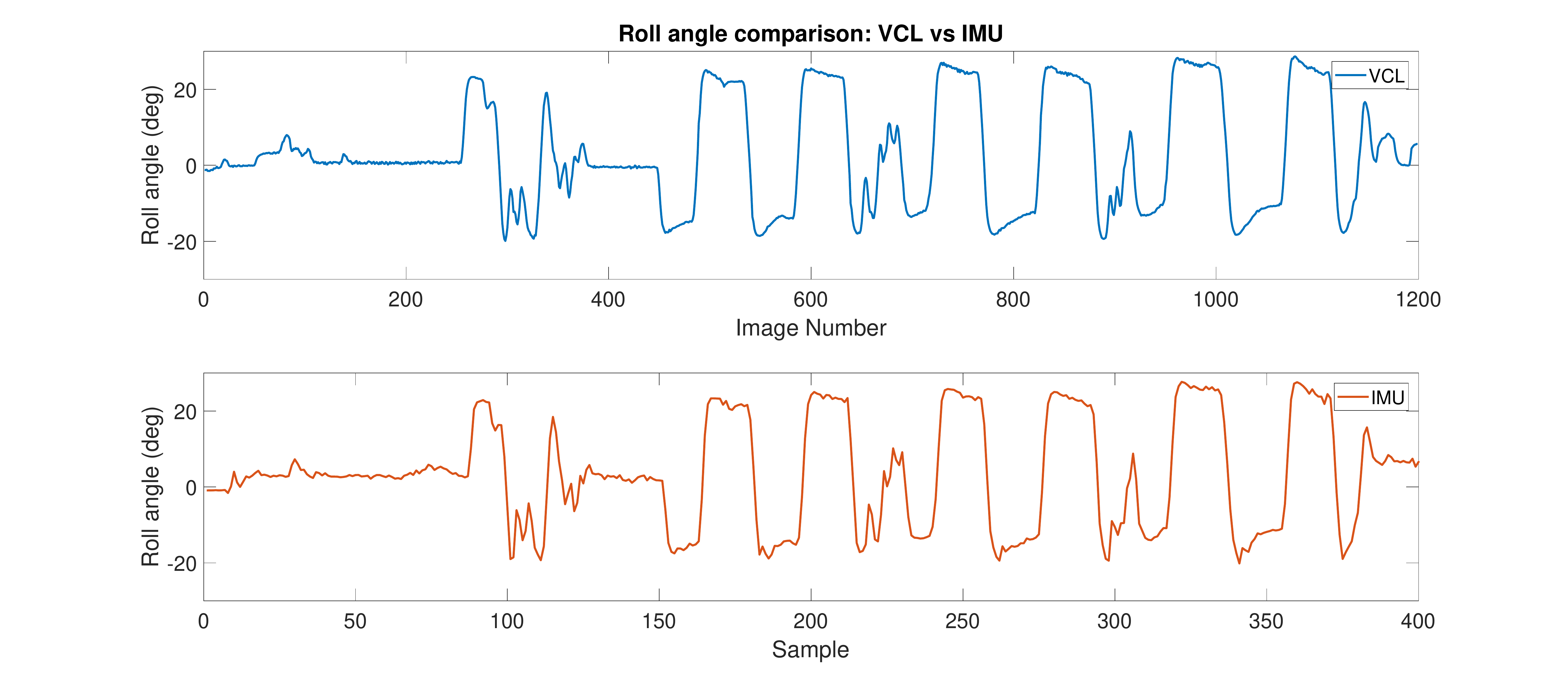}
	\caption{Roll angle comparison between VCL estimates and IMU for a fast side-to-side flight.}
	\label{fig:rolltest}
\end{figure}

\begin{figure*}
	\centering
	\captionsetup[subfigure]{justification=centering}
	\subfloat[V1 with no fusion]{{ \includegraphics[width = 3.2cm,height=3.5cm]{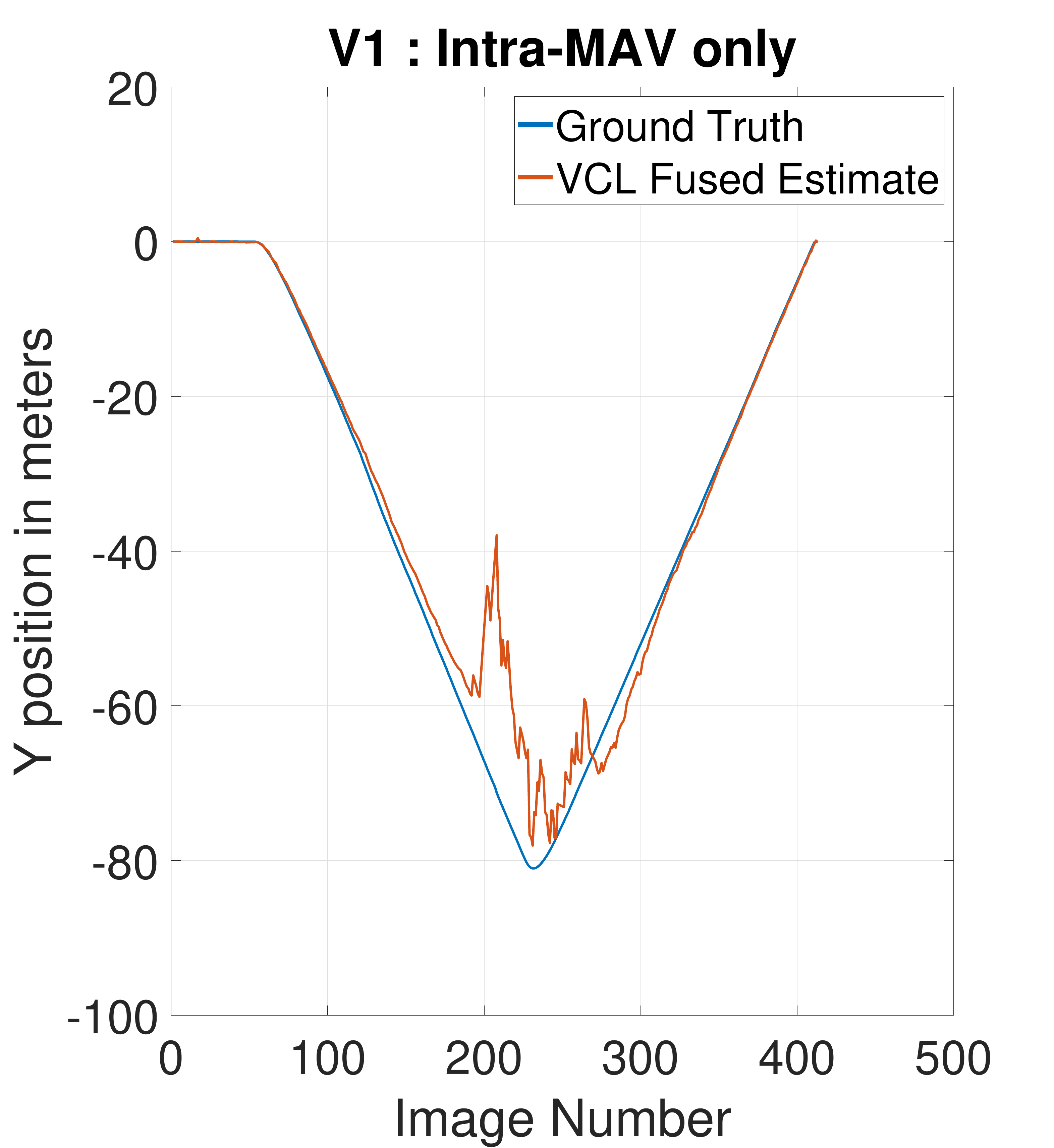} }} %
	\hspace{0.1em}
	\subfloat[V1 with fusion every 50 images]{{ \includegraphics[width = 3.2cm,height=3.5cm]{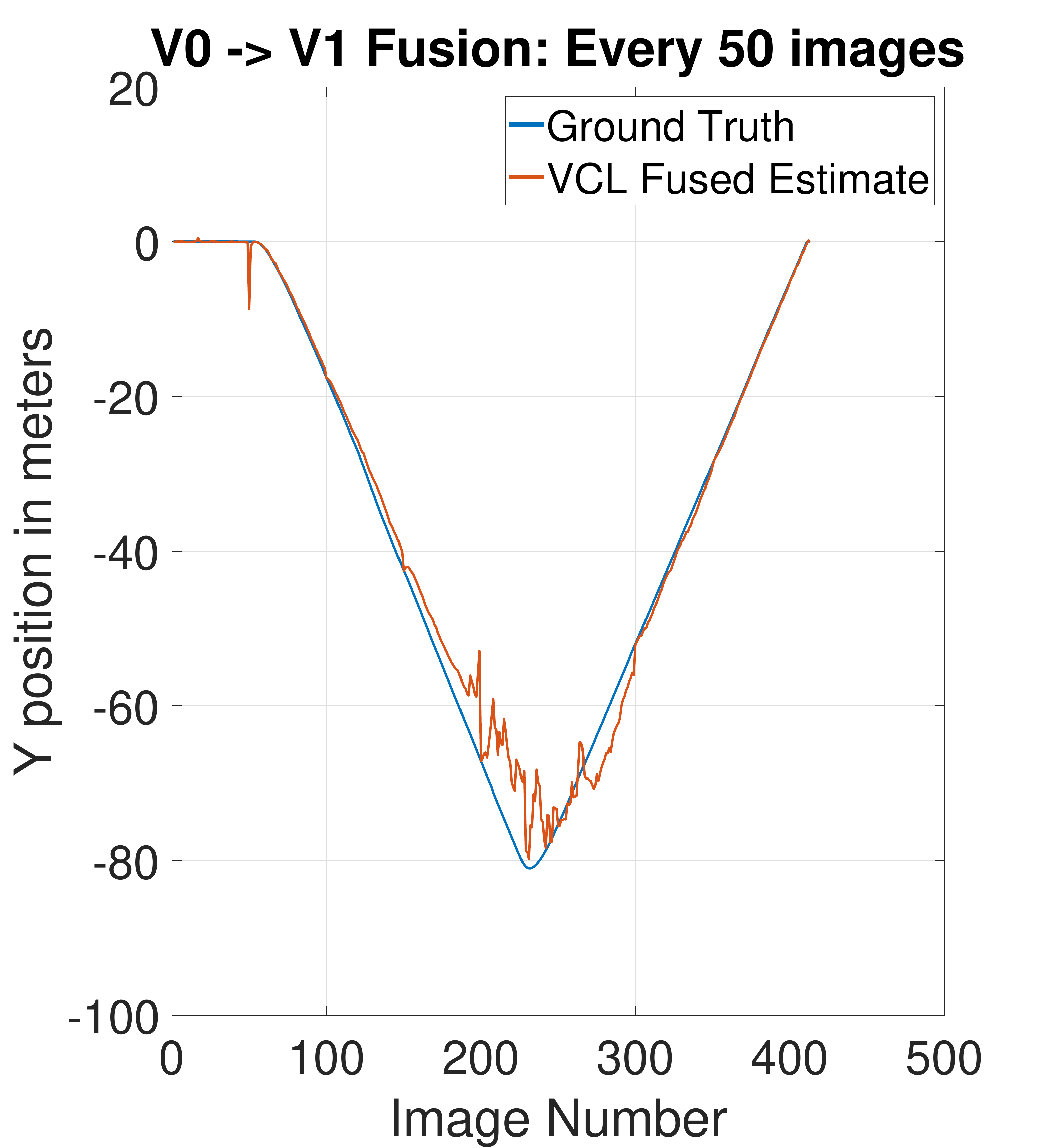} }} %
	\hspace{0.1em}
	\subfloat[V1 with fusion every 20 images]{{ \includegraphics[width = 3.2cm,height=3.5cm]{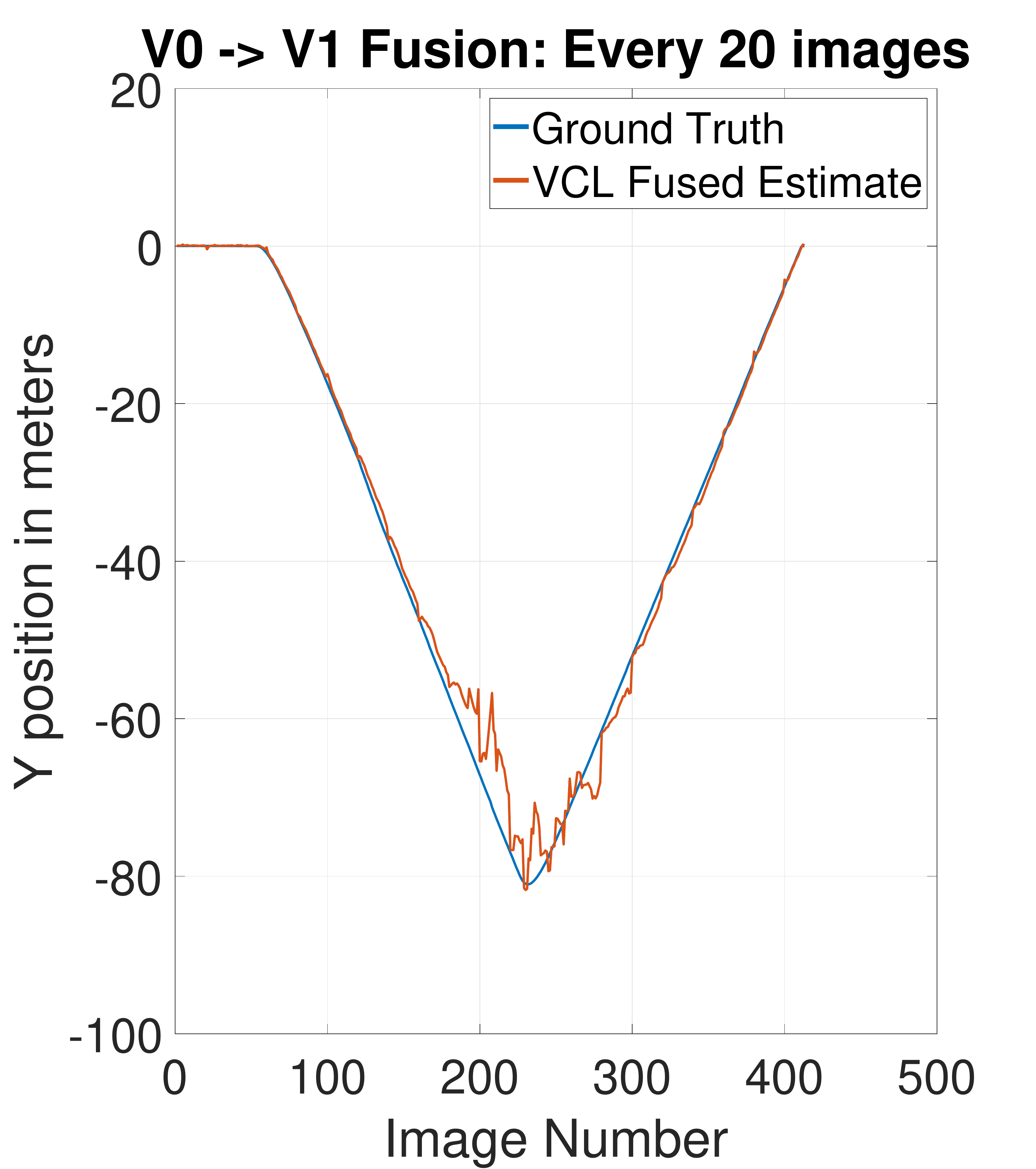} }} %
	\hspace{0.1em}
	\subfloat[V1 with fusion every 5 images]{{ \includegraphics[width = 3.2cm,height=3.5cm]{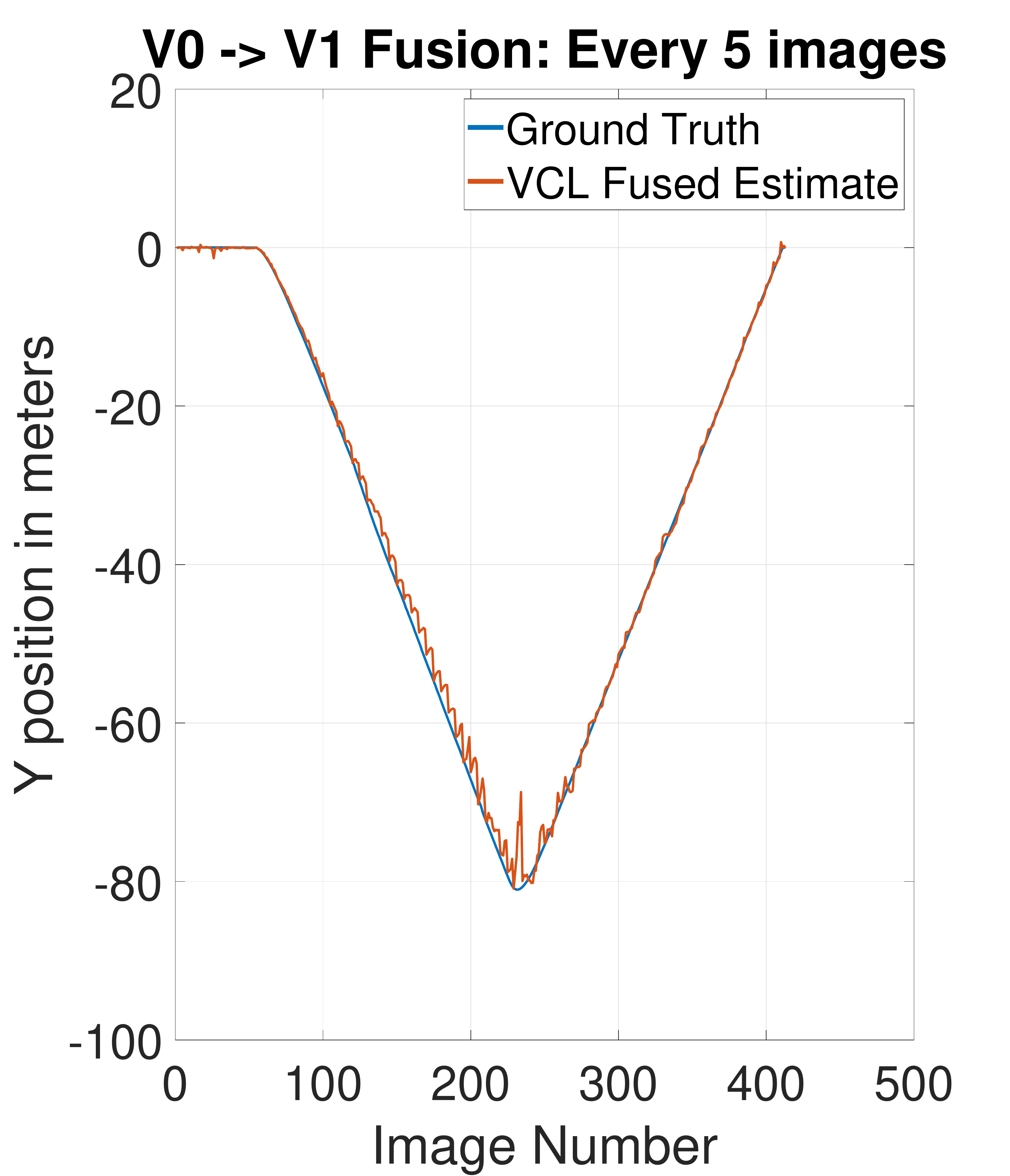} }}
	\hspace{0.1em}
	\\
	\subfloat[V2 with no fusion]{{ \includegraphics[width = 3.2cm,height=3.5cm]{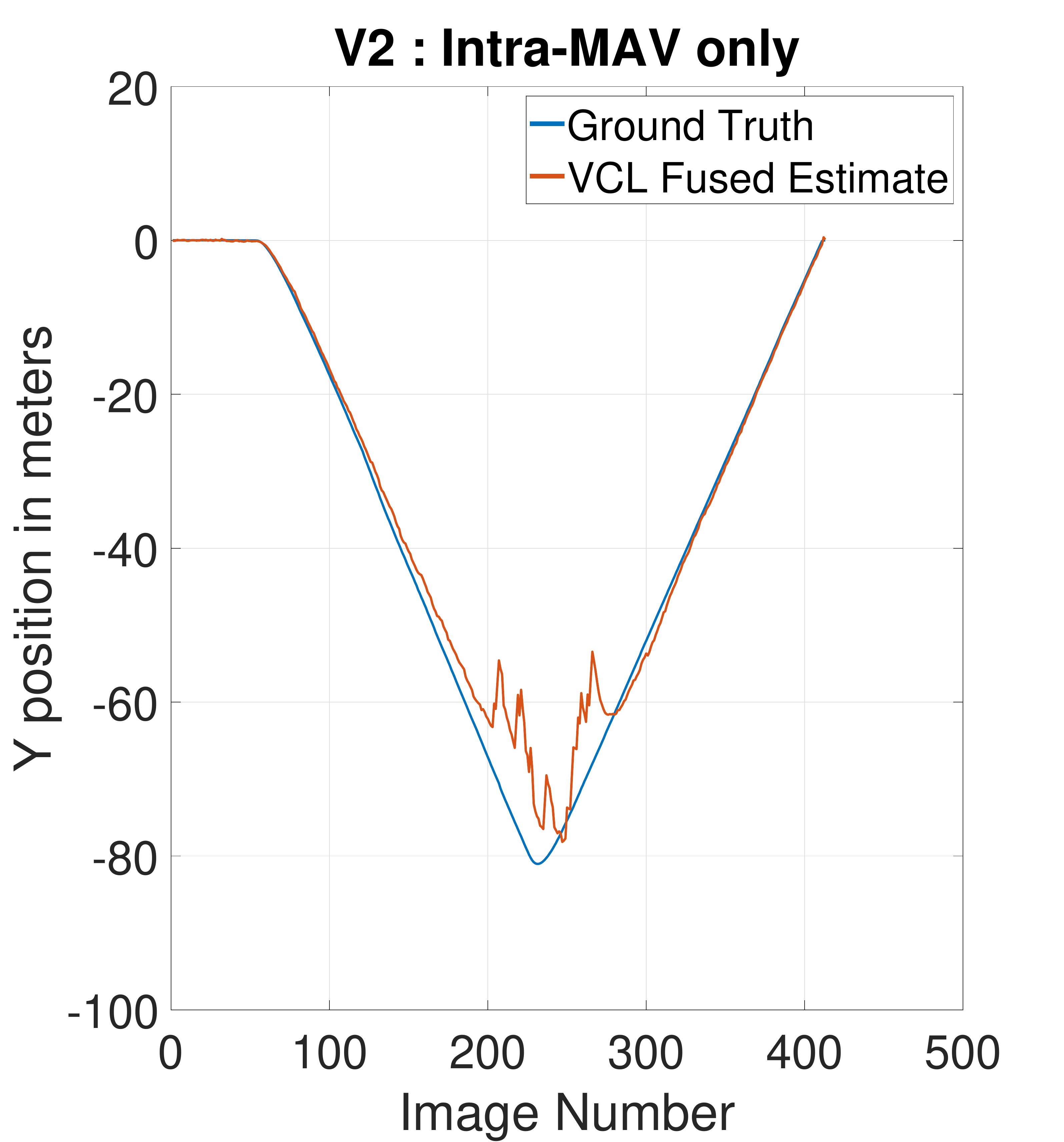} }} %
	\hspace{0.1em}
	\subfloat[V1 with fusion every 50 images]{{ \includegraphics[width = 3.2cm,height=3.5cm]{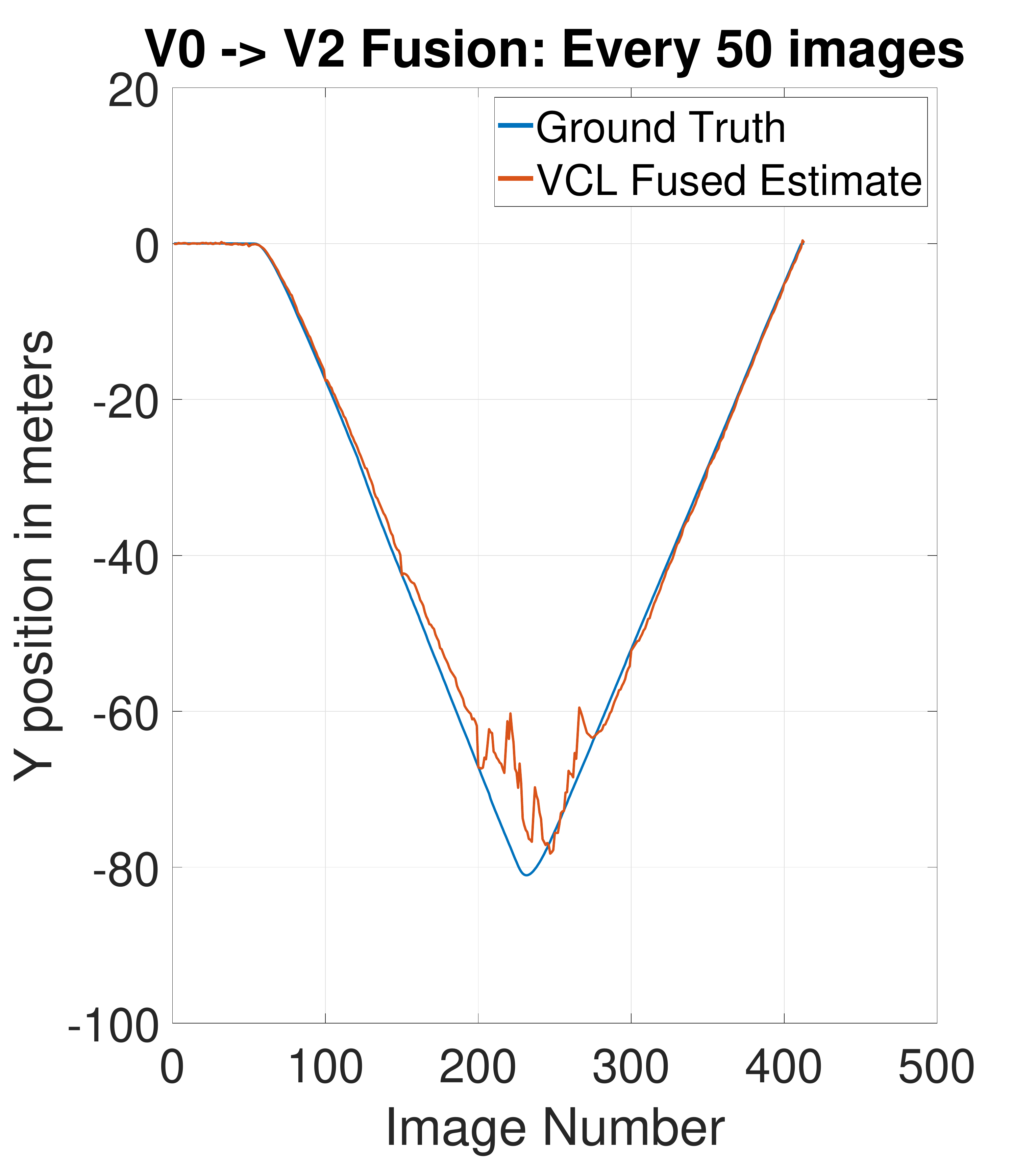} }} %
	\hspace{0.1em}
	\subfloat[V2 with fusion every 20 images]{{ \includegraphics[width = 3.2cm,height=3.5cm]{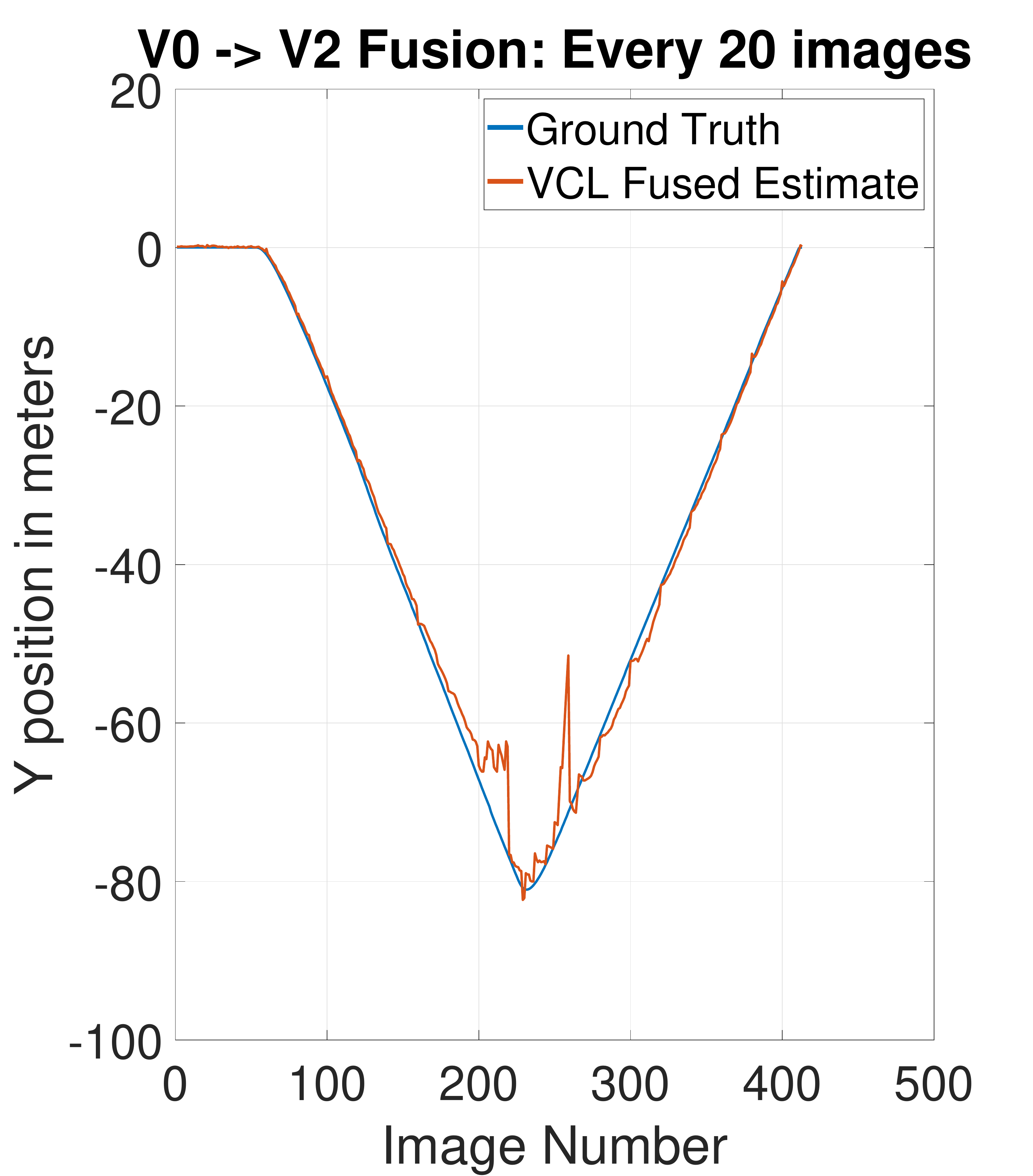} }} 	
	\hspace{0.1em}
	\subfloat[V2 with fusion every 5 images]{{ \includegraphics[width = 3.2cm,height=3.5cm]{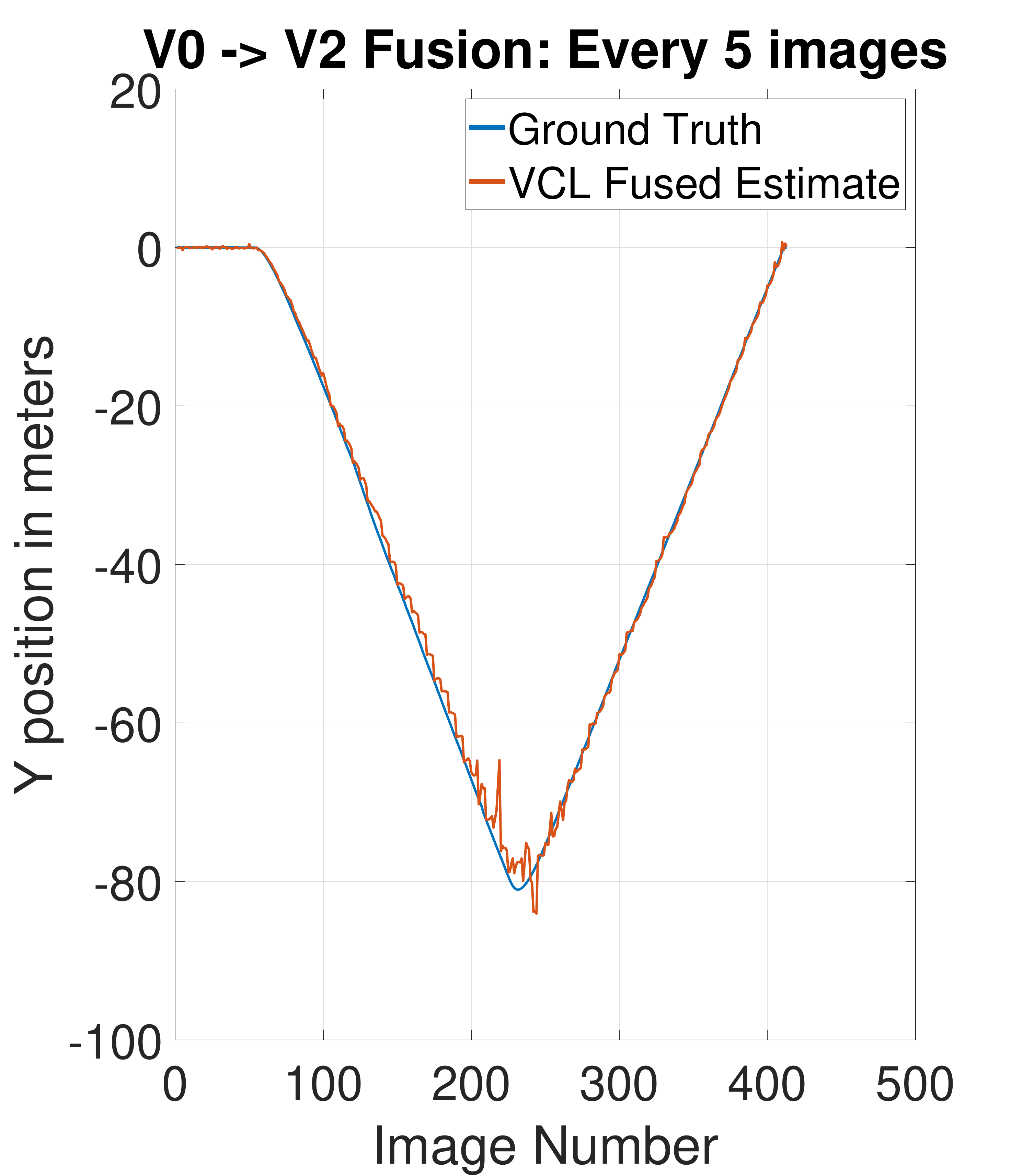} }}
	\caption{Effect of frequency of inter-MAV data fusion on localization for both clients: Only Y axis positions shown}
	\label{fig:intertimemse}
\end{figure*}

\subsection{Intra-MAV localization: real experiments}
Similar intra-MAV localization was also tested in real scenarios with the Bebop 2 quadrotors. At first, the vehicles collaborate to isolate common features and build a map, within which each MAV attempts to localize itself. Figure \ref{fig:intrareal} show the localization results for rectangle shaped trajectories navigated by two MAVs. The first test (\ref{fig:intrareal}(a)) was meant to evaluate the accuracy of the algorithm against ground truth, so the MAVs were moved by hand along two pre-marked rectangles of dimensions 75$\times$20 feet each. The VCL estimates were seen to track the ground truth fairly well, while exhibiting a slight drift at the far edges (which can be attributed to changes in feature appearance when at a distance of 75 feet from the initial location where the map was built).

The second test of the intra-MAV localization involved manual flight of two MAVs in an outdoor area near a building, and comparison with the GPS position estimates. In this test, it was observed that the VCL estimates were closer to the real trajectories taken by the MAVs than the GPS, which could be attributed to low altitudes and thereby a certain degree of inaccuracy in the GPS. The results of this test can be seen in \ref{fig:intrareal}(b).

We have also tested the accuracy of the orientation estimates in real flights. In one particular test, the two Bebop MAVs were made to fly side to side at high speeds (up to 5 m/s), and the estimates of the roll angles were compared to the estimates coming from the IMU data onboard the vehicle. The VCL estimates and the IMU estimates of the angles are close to each other (Figure \ref{fig:rolltest}), demonstrating an accurate estimation of angles even through fast flights. The roll angles coming from the IMU and the VCL are plotted separately instead of a common X axis due to the varying frequencies of the data sources.

\subsection{Inter-MAV localization: simulation}

\subsubsection{Effect of frequency of relative measurements}
\begin{figure}
	\centering
	\captionsetup[subfigure]{justification=centering}
	\subfloat[Initial position]{{\includegraphics[width = 2.5cm,height=2cm]{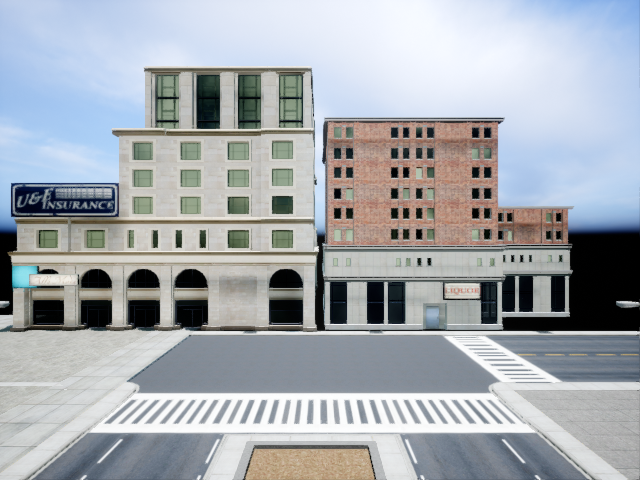}}}
	\quad
	\subfloat[Midpoint]{{\includegraphics[width = 2.5cm,height=2cm]{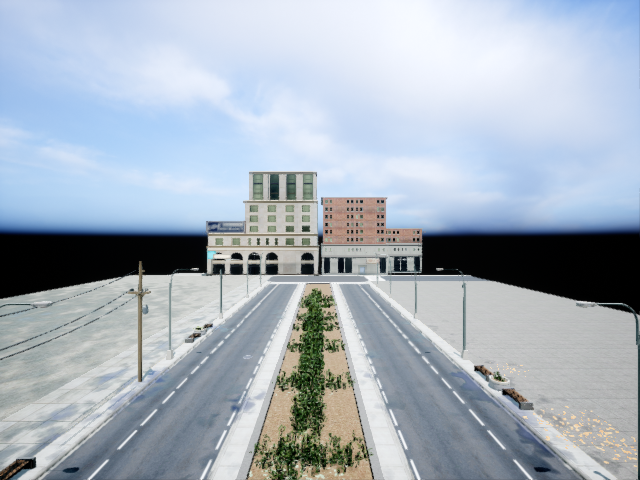}}}
	\quad
	\subfloat[Final position]{{\includegraphics[width = 2.5cm,height=2cm]{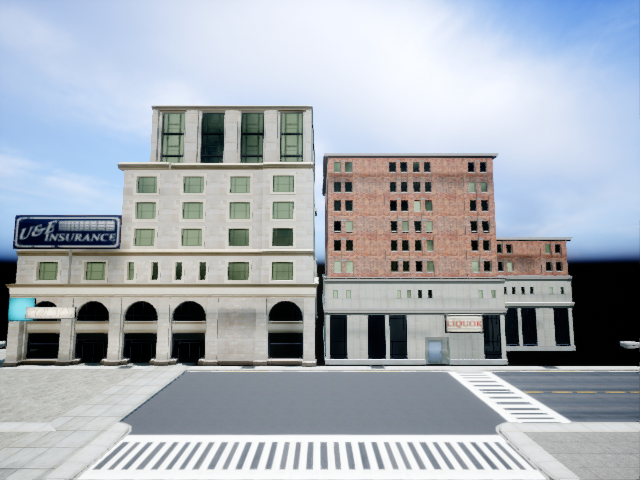}}}
	\caption{Inter-MAV estimation test: backward and forward trajectory sample images}
	\label{fig:interexample}
\end{figure}

Inter-MAV localization was first tested and evaluated in the simulation, by creating a sparsely populated environment where three vehicles start off side by side in front of a building, but are then commanded to fly backwards (thus away from features) and then forwards to the starting positions. Three images from this trajectory are shown in Figure \ref{fig:interexample} to demonstrate the change in feature appearances. Between the backward and forward motions, each MAV traverses 80 + 80 = 160 meters in the environment.  

\begin{figure}
	\centering
	\includegraphics[width = 8cm]{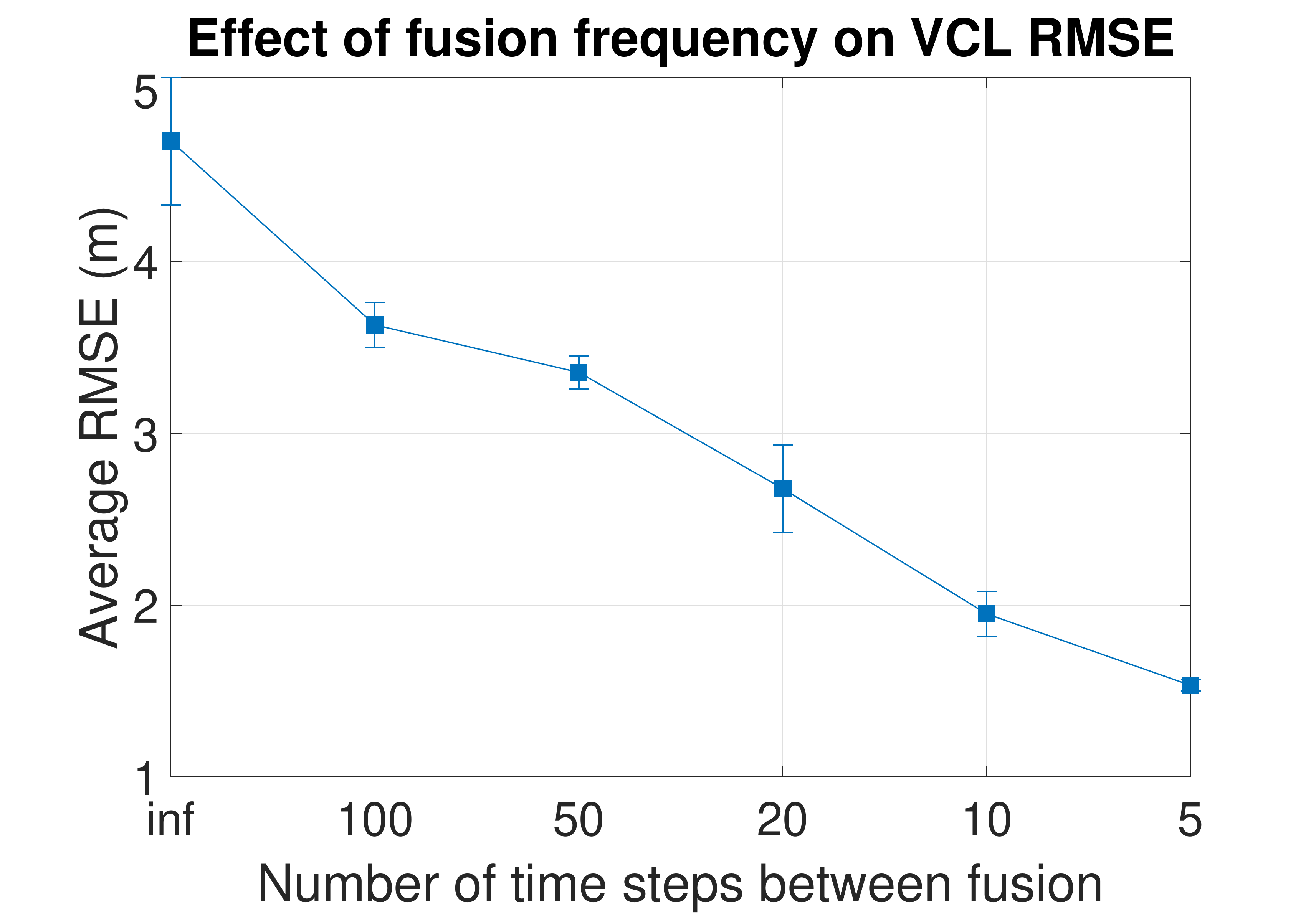}
	\caption{Effect of frequency of inter-MAV data fusion on RMS error. Error bar shows variation of position RMSE between clients V1 and V2. }
	\label{fig:interfreqmse}
\end{figure}

\begin{figure*}
	\centering
	\subfloat[One host]{{ \includegraphics[width=0.24\textwidth]{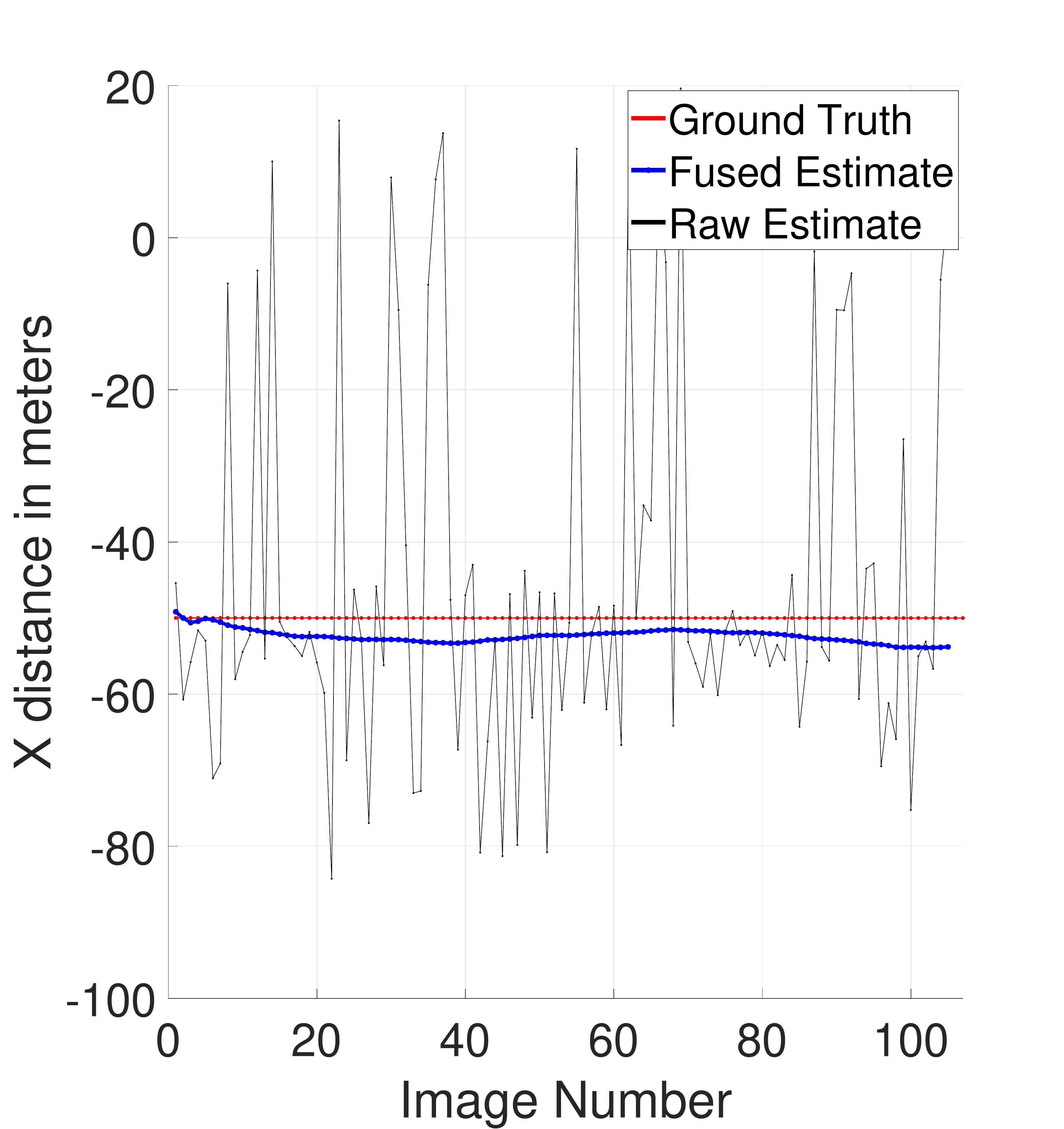} }} 
	\subfloat[Two hosts]{{ \includegraphics[width=0.24\textwidth]{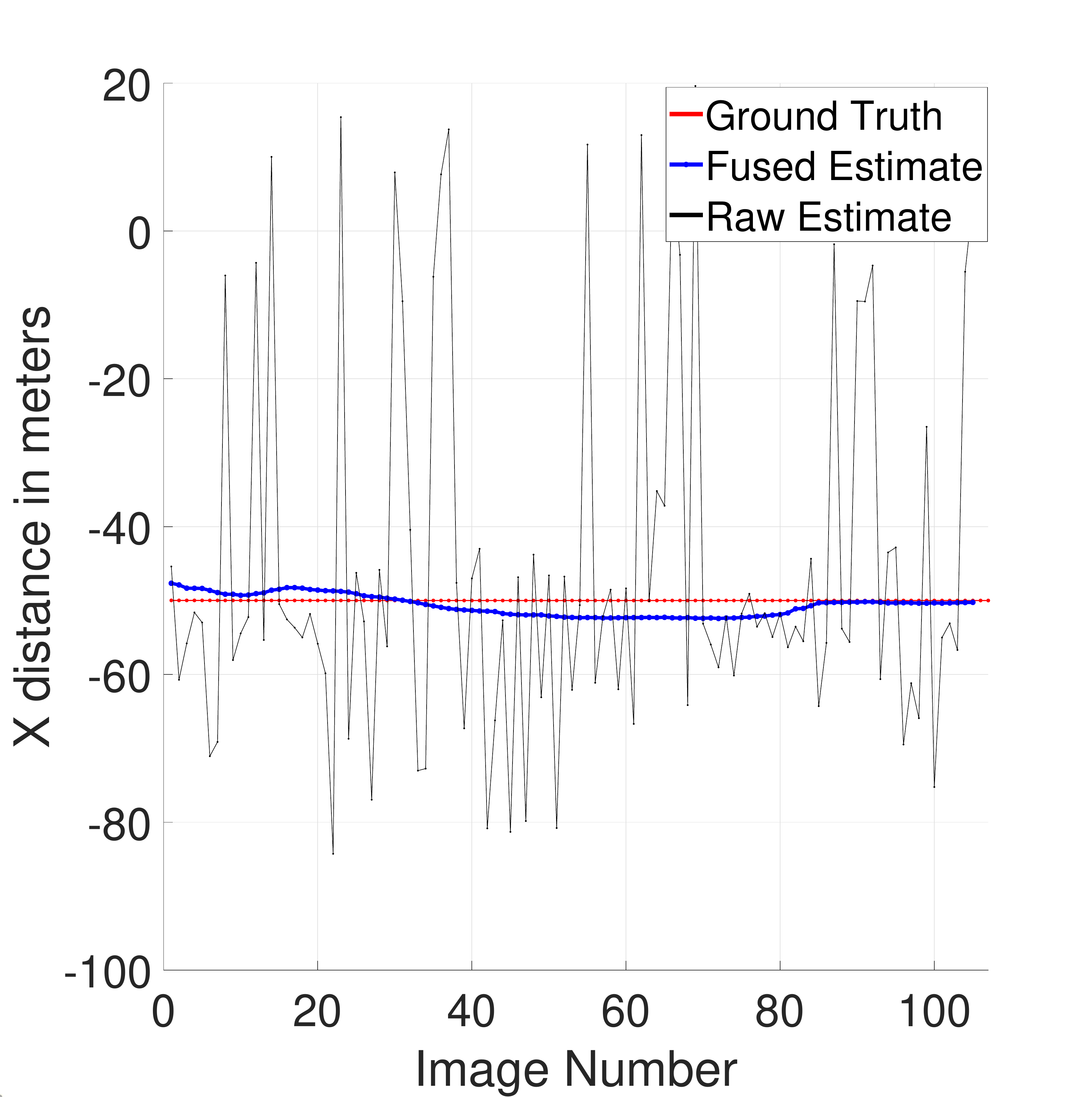} }}
	\subfloat[Three hosts]{{ \includegraphics[width=0.24\textwidth]{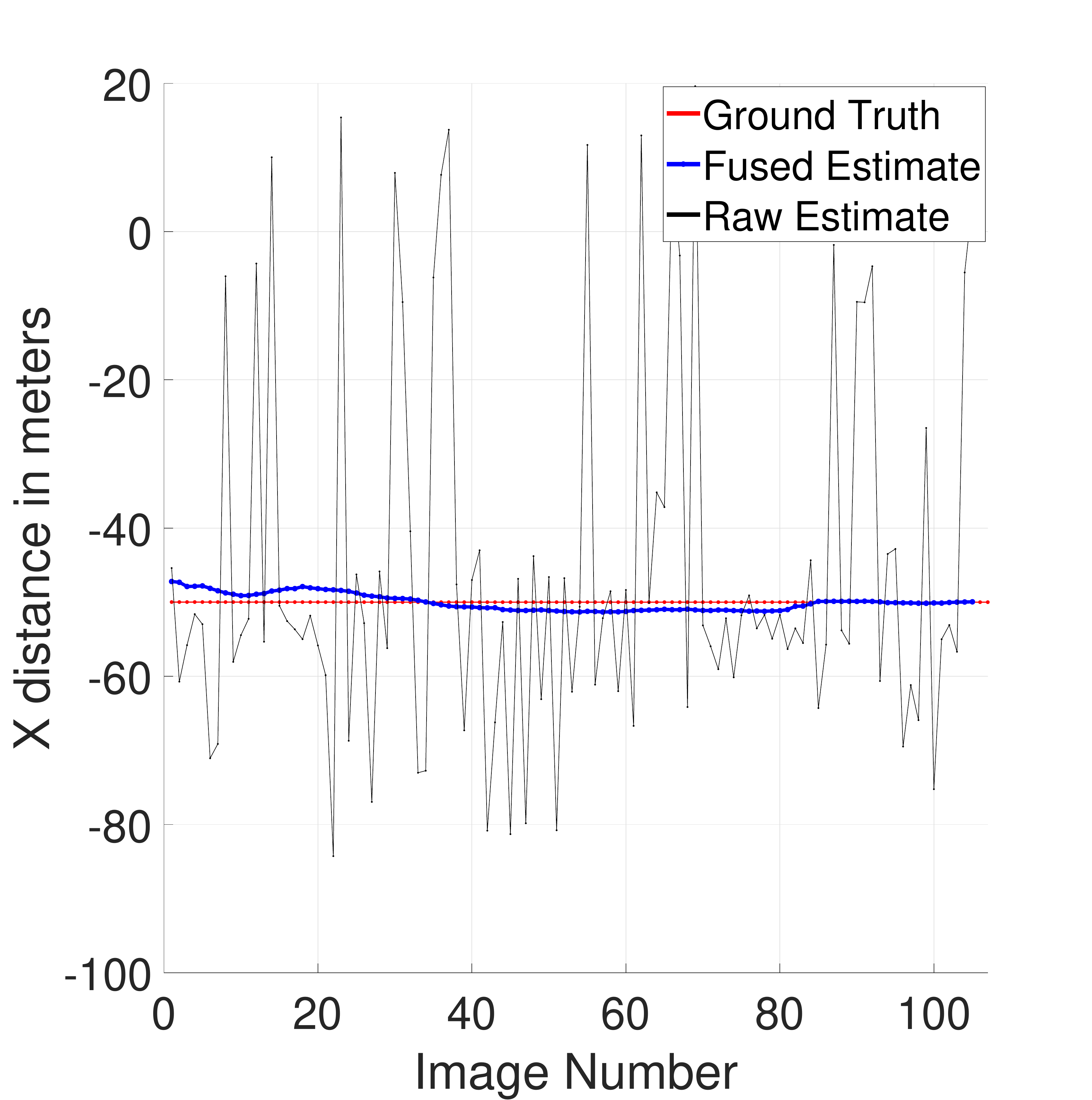} }}
	\caption{Client X axis position estimate with varying number of sources for fusion}
	\label{fig:fusioncompvehicle}
\end{figure*}
We recall here that the biggest advantage of inter-MAV localization is when one vehicle has better localization than the others, so to simulate such a condition, we assume the MAV in the center has access to better estimates. We use the ground truth values of the positions for the MAV in the center and corrupt them with zero-mean noise in order to simulate 'good localization', and then attempt to analyze the effect of inter-MAV localization between this host MAV and the others. Alongside, intra-MAV localization was performed at every time step for both the client MAVs ousing the captured images.

With no inter-MAV measurements, the VCL estimates for the clients show significant error at the midpoint, where the vehicles are the farthest from the scene. But once inter-MAV measurements are obtained and fused, the errors decrease, and increasing the number of times inter-MAV measurements are fused also decreases the error significantly (Figures \ref{fig:intertimemse} and \ref{fig:interfreqmse}). In this specific case, increasing the frequency of the relative measurements to once in 5 images brings the fused position estimates of $V_1$ and $V_2$ closer to the ground truth, exhibiting a RMS error of 2.3m in position. Over 160m of navigation, this equates to less than 1\% of RMS error.

\begin{figure}
	\centering
	\includegraphics[width = 6cm,height=4cm]{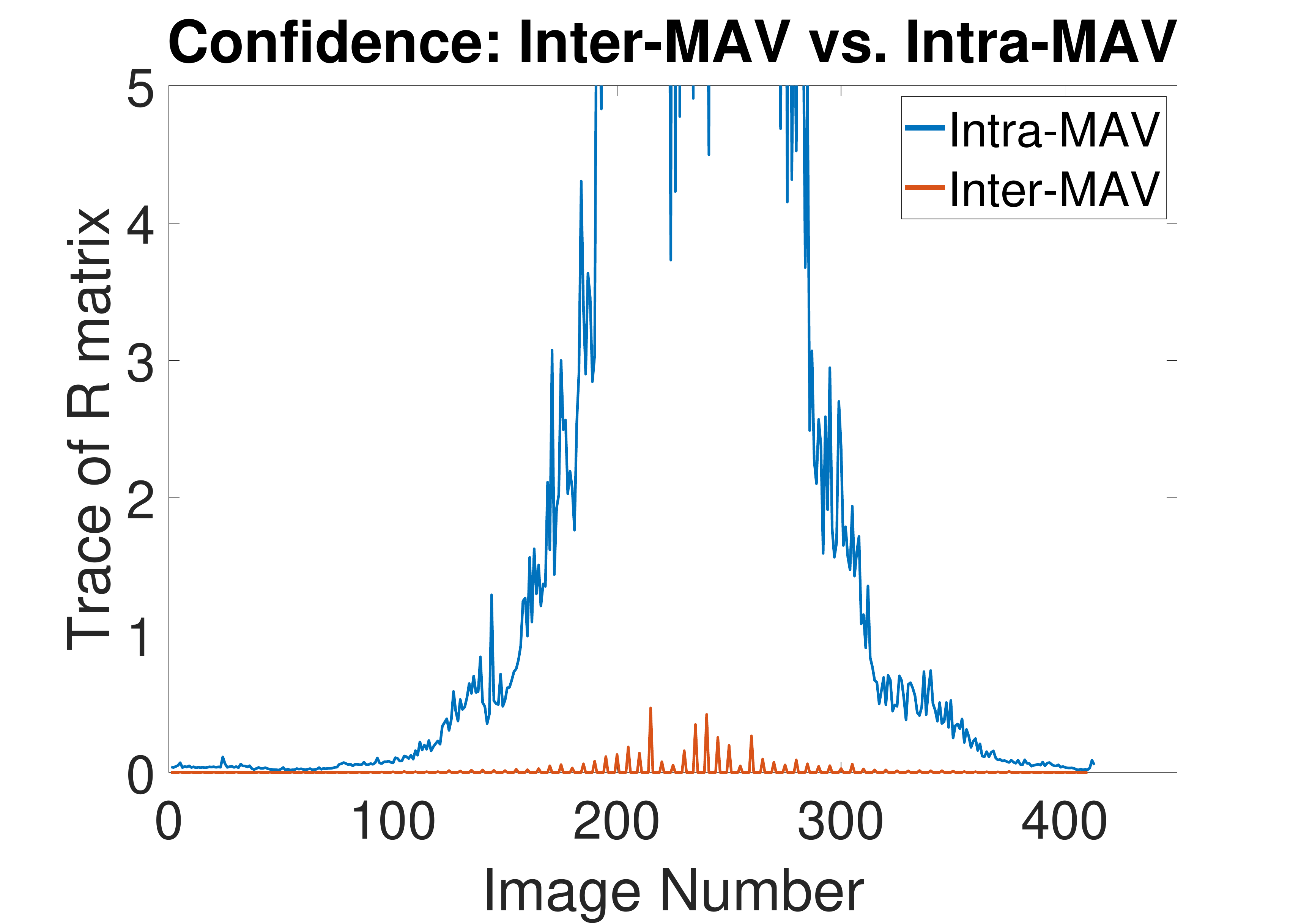}
	\caption{Comparison of inter vs intra-MAV measurement covariances. Inter-MAV measurements are usually seen to have significantly lower solution covariance.}
	\label{fig:intercovcomp}
\end{figure}

Figure \ref{fig:intercovcomp} shows a comparison of the measurement confidences of intra-MAV measurements versus inter-MAV measurements by plotting the trace of the measurement covariance $\mathbf{R}$. It is evident here that as the MAV moves away from the map features, the feature appearance deviates from the keyframes and the intra-MAV measurement covariance rises rapidly; but the inter-MAV covariance stays relatively low throughout, as it only depends on the amount of feature overlap and distribution of points at every particular instant. 

\subsubsection{Effect of number of vehicles in group}
\begin{figure}
	\centering
	\subfloat[$V_0$]{{ \includegraphics[width=0.1\textwidth]{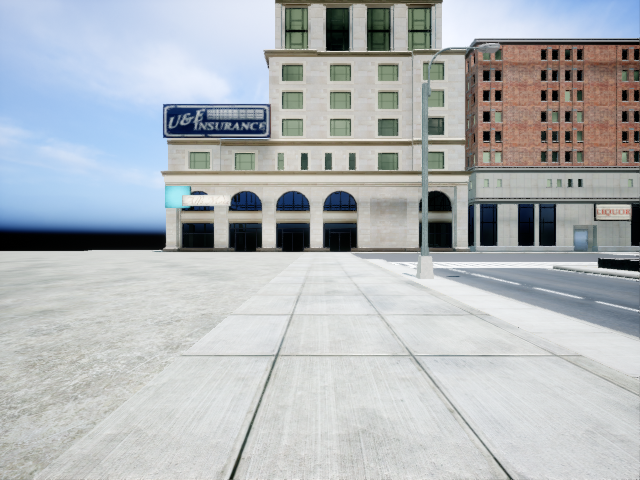} }} 
	\subfloat[$V_1$]{{ \includegraphics[width=0.1\textwidth]{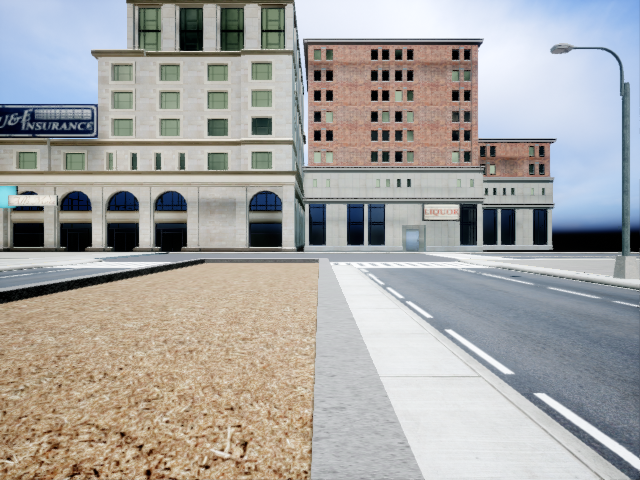} }}
	\subfloat[$V_2$]{{ \includegraphics[width=0.1\textwidth]{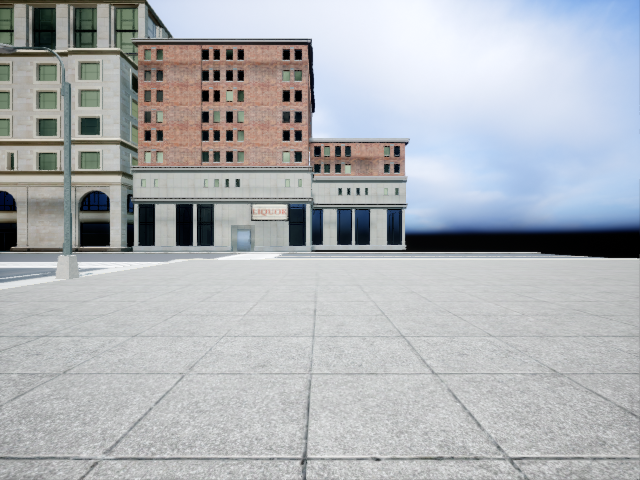} }}
	\subfloat[$V_3$]{{ \includegraphics[width=0.1\textwidth]{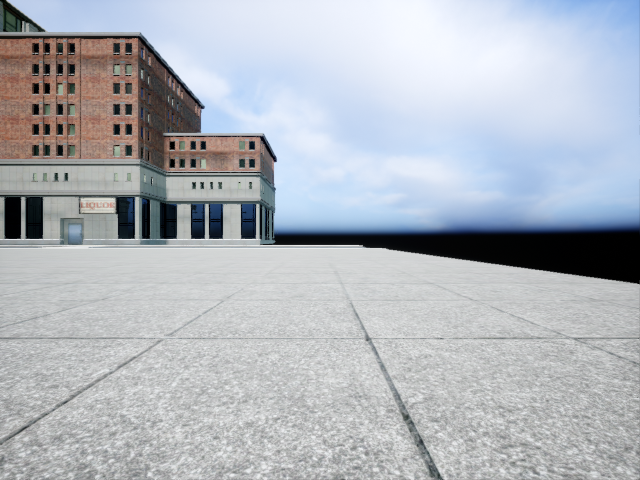} }}
	\caption{Images from the sarting positions for four MAVs in AirSim}
	\label{fig:inter5}
\end{figure}

In a second experiment involving inter-MAV localization and fusion, we attempt to evaluate the effect of having more than one host vehicle that is able to assist with inter-MAV localization for a client vehicle. For this experiment, we initialize four MAVs in simulation, which are spaced apart by a distance of 25m between each pair of vehicles, which is considerably higher compared to their distance from the scene. We treat $V_3$ as the client that needs localization assistance, because it can be seen from Figure \ref{fig:inter5} that $V_3$ has the least overlap with the scene, thus has the most potential to exhibit inaccuracy in the intra-MAV localization scheme. The vehicles were then made to fly vertically up and down, hence, their X axis position variation was minimal (to demonstrate the problem with scene overlap). 

\begin{figure}
	\centering
	\includegraphics[width = 9cm]{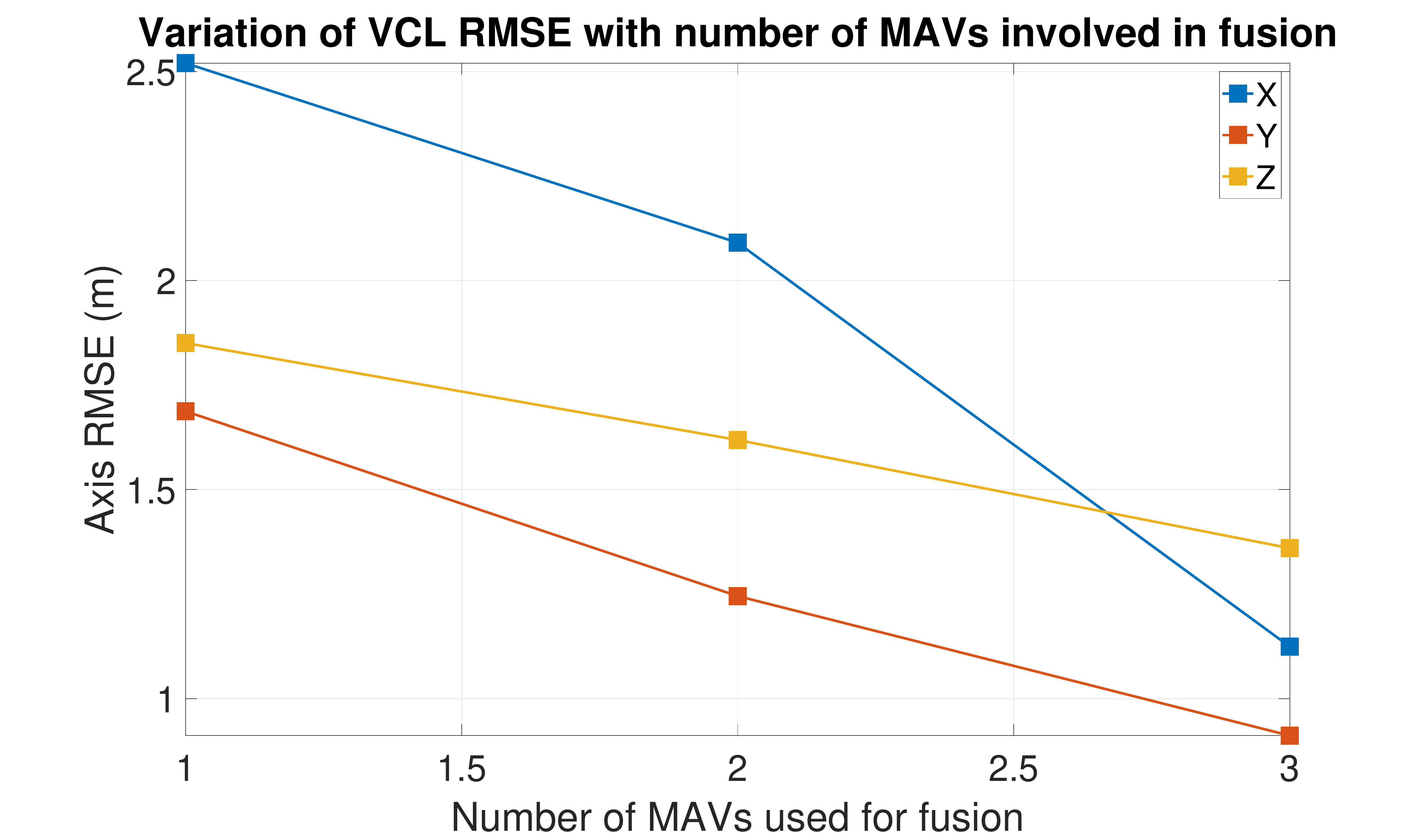}
	\caption{Change in position RMSE of one MAV with the number of participants providing relative measurements}
	\label{fig:inter5mse}
\end{figure}

\begin{figure*}
	\centering
	\captionsetup[subfigure]{justification=centering}
	\subfloat[Intra-MAV only localization]{{ \includegraphics[width = 8cm,height=5cm]{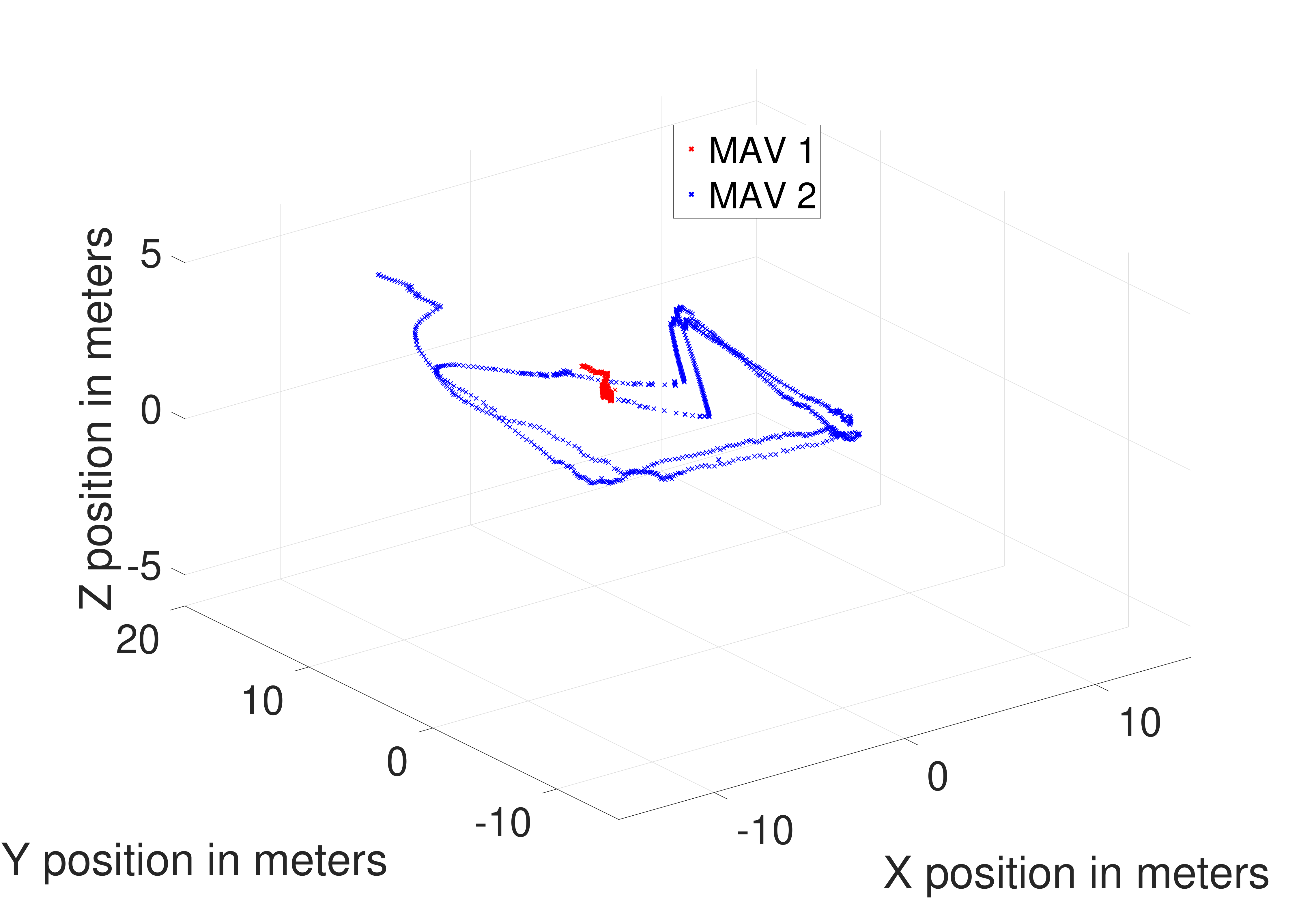} }} 
	\subfloat[Inter+intra fused localization]{{ \includegraphics[width = 8cm,height=5cm]{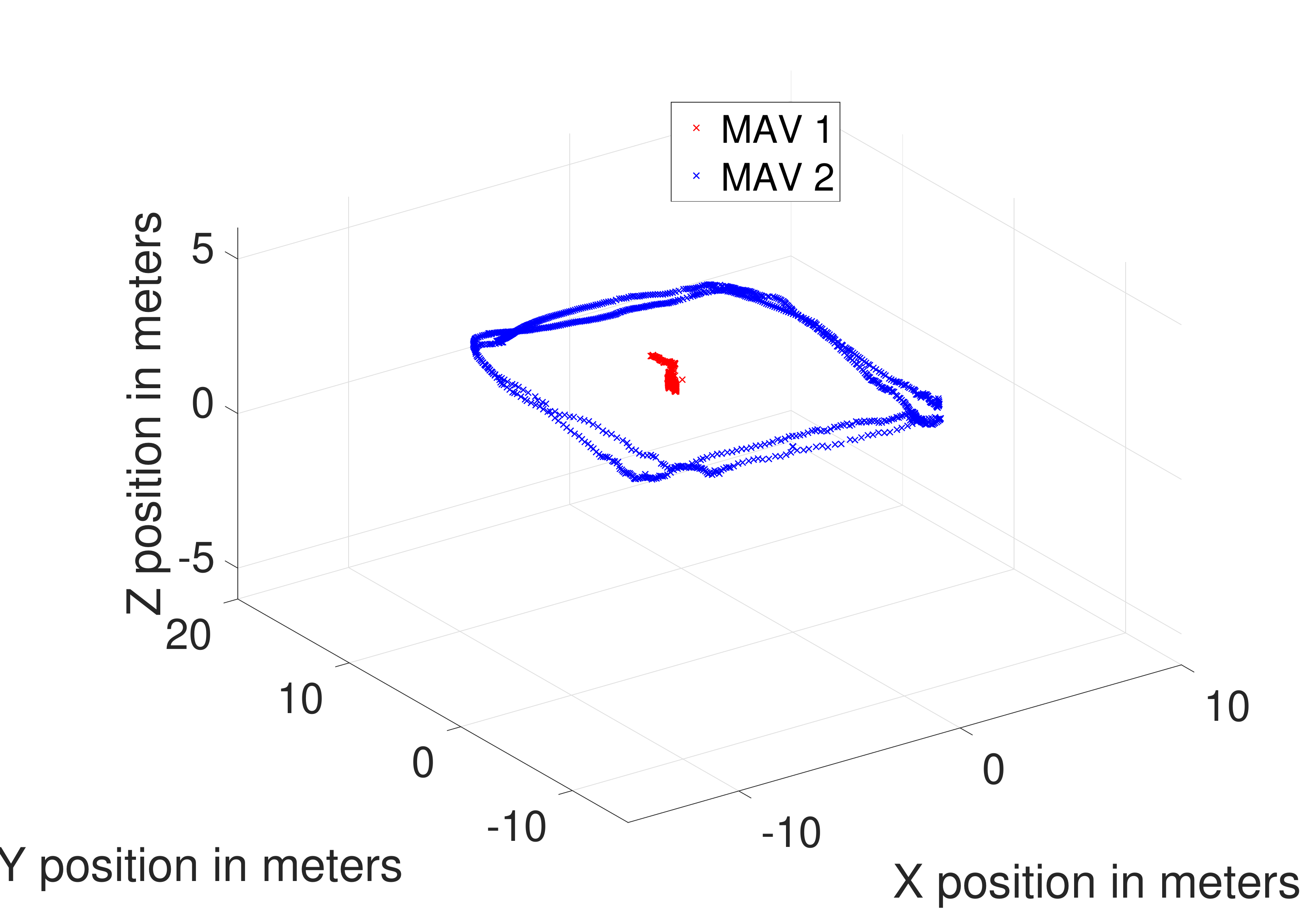} }} 
	\caption{Indoor flight with two MAVs: intra-only localization vs. inter+intra fused localization.}
	\label{fig:indoorinter}
\end{figure*}

We observe that all three hosts combining their inter-MAV estimates results in the most accurate estimate for the client $V_3$, as multiple sources of information help robustify the final estimate in the covariance intersection scheme. In Figure \ref{fig:fusioncompvehicle}, we show the X-axis position estimates of $V_3$ compared to the ground truth, with the fused estimate in blue and the raw estimate in dotted black. Although $V_3$'s internal estimate is very noisy, fusing data from one other `host' , as seen in \ref{fig:fusioncompvehicle}(a) allows $V_3$ to track the ground truth with an RMSE of 2.52 m. The error reduces even further with an increase in the number of vehicles taking part in the fusion, as evidenced by the estimates in figures \ref{fig:fusioncompvehicle}(b) and \ref{fig:fusioncompvehicle}(c), with three hosts bringing the error down to 1.12 m. Figure \ref{fig:inter5mse} plots the variation of the RMSE on all three position axes as the number of vehicles participating in data fusion changes.

\subsubsection{Handling pure rotation}
\begin{figure}
	\centering
	\begin{tabular}{cc}
		\subfloat[Fused yaw angle estimate for a rotating MAV.]{\includegraphics[width=4cm, height=3.5cm]{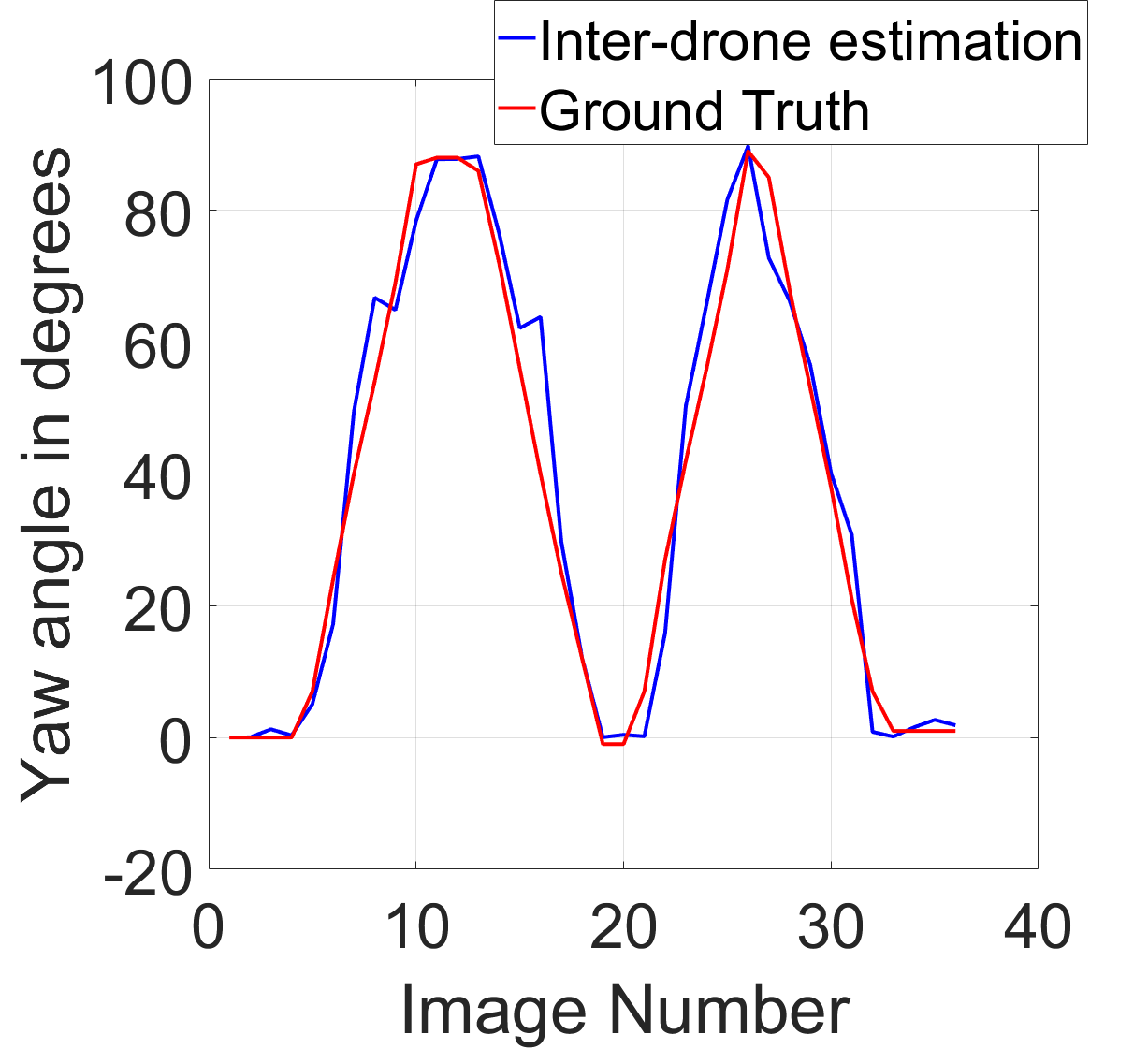}} &
		\subfloat[Yaw angle from relative estimates from the other two vehicles, and associated reprojection errors.]{\includegraphics[width=3.7cm, height=3.7cm]{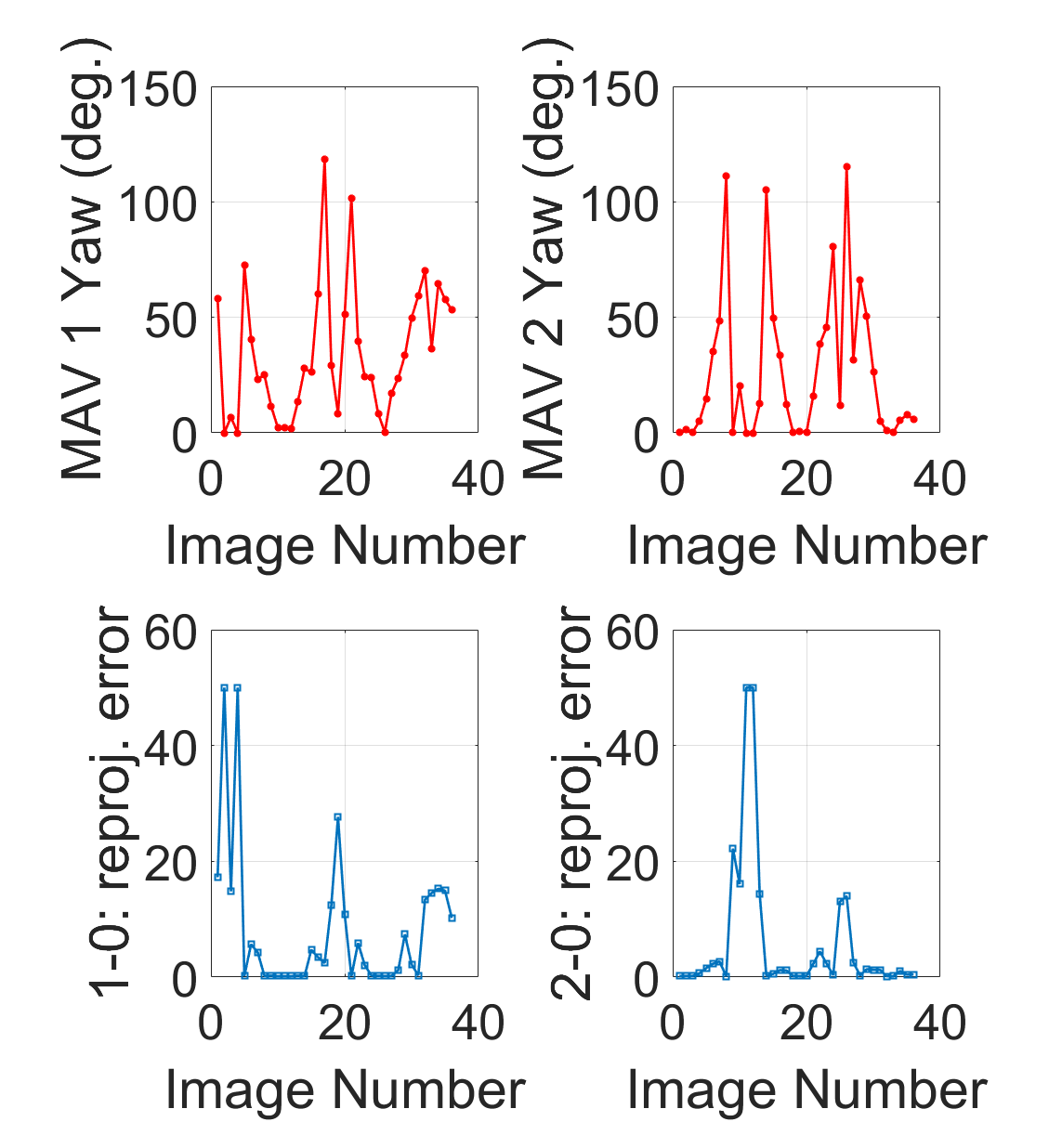}} \\
	\end{tabular}
	\caption{Estimation of yaw angle of a client MAV through inter-MAV localization when other MAVs are able to contribute.}
	\label{fig:yawinter}
\end{figure}
One of the problems that is evident in single monocular-camera localization is the issue with purely rotational movement in yaw, which is a very common maneuver for MAVs. Pure rotation usually causes a large portion of existing map points to go out of view suddenly, while the fact that there is no translation by the camera means it is not possible to triangulate new feature points through a single camera without any additional information. In contrast, the VCL algorithm is able to handle this problem by capitalizing upon the inter-MAV localization feature points between what is one rotating MAV and another MAV that has map points in common.

As a test case, we simulated an environment containing two perpendicular buildings being observed by an MAV (MAV 0) that performs periodic 90-degree rotations, trying to observe both. MAV 0's rotation speed was set to 45$^\circ$/s. Two other MAVs (1 and 2) are also present in the proximity in hover mode, each observing one of the buildings from a distance. To assist with MAV 0's pose estimation during fast rotations, for every image captured by these three MAVs, relative poses are computed between vehicles 1-0 and 2-0 and fusion is attempted with the estimate onboard MAV 0 itself. Due to the way the fusion algorithm is framed, the final fused yaw angle estimate is computed based on which estimate has the least uncertainty. In figure \ref{fig:yawinter}(a), the fused estimates of the yaw angle are shown, where it can be seen that the estimated angle closely matches the ground truth. A careful analysis of figure \ref{fig:yawinter}(b) shows how either MAV 1 or MAV 2 is chosen as the better source of the yaw information based on the reprojection error (on which the uncertainty estimate depends). Swift rotations such as this are typically problematic for single-camera localization, but the collaborative aspect can maintain localization through data fusion from other, possibly more reliable, sources of information.

\subsection{Inter-MAV localization: real experiments}

After validating the efficacy of inter-MAV localization and data fusion in simulation, we tested the VCL algorithm with fusion on data from real flights. As a first experiment, two Bebop MAVs were flown indoors, where both MAVs first obtained initial images to map a room, and then one MAV (MAV 1) was commanded to hover in the middle of the room, whereas the other (MAV 2) navigated a square trajectory around the first. This trajectory was traversed in a way that MAV 2 encounters one part of the room where the feature overlap with the original map is relatively low. As a result of this, relying solely on intra-MAV localization resulted in a high amount of drift and measurement outliers in that particular part of the environment, creating an inaccurate trajectory that can be seen in Figure \ref{fig:indoorinter}(a).

To alleviate this problem, the VCL algorithm was made to perform inter-MAV localization between MAVs 1 and 2, whenever the number of features tracked by MAV 2 fell under a threshold. The covariance of inter-MAV localization being much lower, it took over whenever intra-MAV localization suffered in accuracy, and the fusion resulted in a significantly more accurate trajectory estimate, as shown in Figure \ref{fig:indoorinter}(b).

\subsection{Map updates}

\begin{figure}
	\centering
	\captionsetup[subfigure]{justification=centering}
	\subfloat[Step 1]{\includegraphics[width = 2.6cm,height=2.6cm]{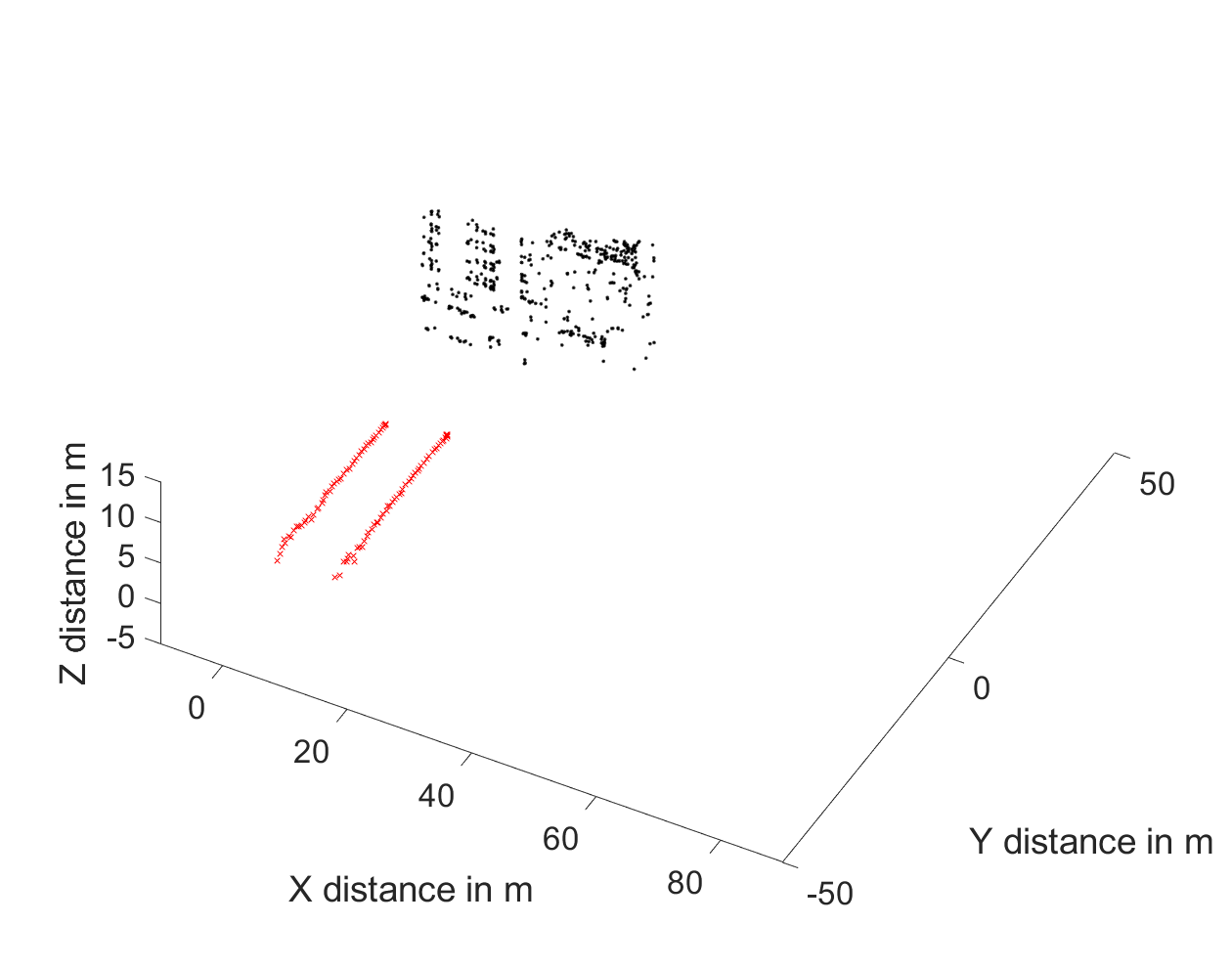}}
	\subfloat[Step 2]{\includegraphics[width = 2.6cm,height=2.6cm]{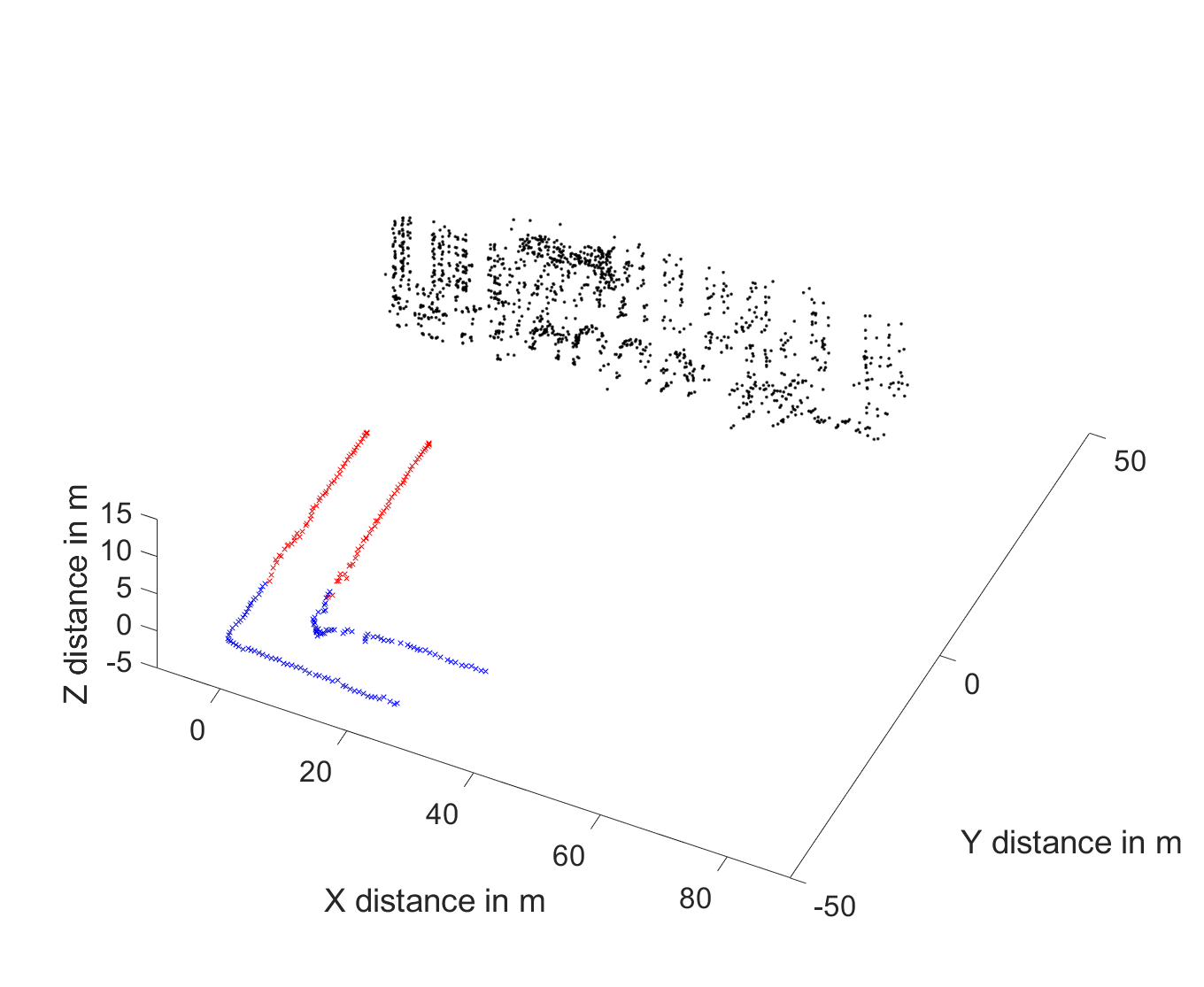}}
	\subfloat[Step 3]{\includegraphics[width = 2.6cm,height=2.6cm]{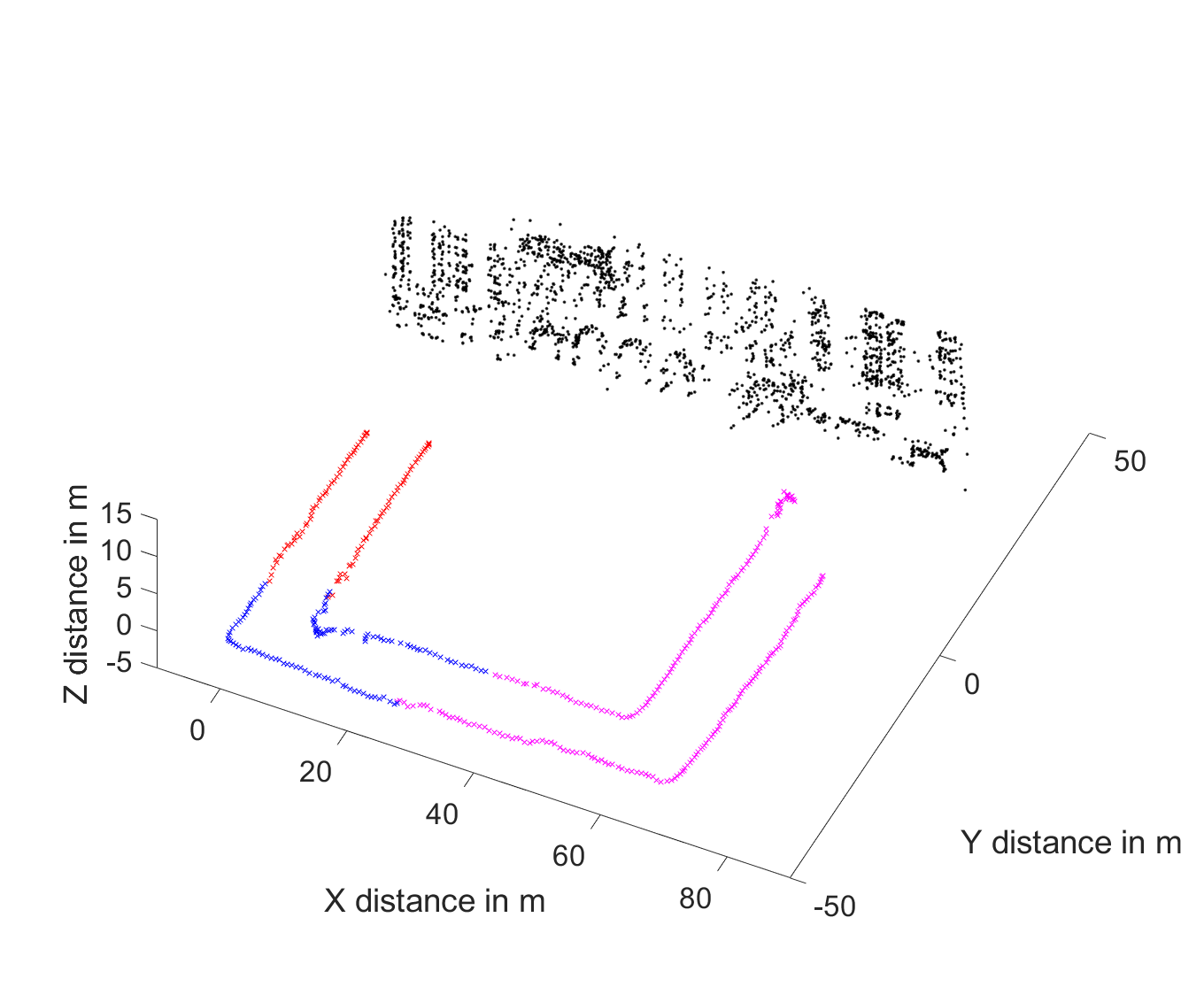}}
	\caption{Process of map updates while maintaining localization: newly captured features are appended to the global map when the number of tracked features becomes low.}
	\label{fig:mapupdates}
\end{figure}

The principles behind relative pose estimation and scale recovery can also be used for performing map updates for the whole group, when the navigation is over large spaces. Figure \ref{fig:mapupdates} shows the process of map updates as two MAVs navigate an environment in AirSim. Table 2 offers some information regarding the accuracy of localization (compared to ground truth) while the vehicles transition through different 3D maps. 

\begin{table}[h]
	\centering
	\begin{tabular}{c | c | c}
		\hline
		MAV  & Total Distance (m) & RMSE error (m)  \\
		\hline
		1 & 170 & 1.35   \\
		2 & 140 & 1.55   \\
		\hline
	\end{tabular}
	\caption[Table caption text]{RMS errors for position estimates of two MAVs localizing through map updates.}
\end{table}

\subsection{Algorithmic requirements and Discussion}

Although the algorithm was not applied in real-time onboard the vehicles, some analysis was performed on the possible computational and communication requirements. It can be recalled that intra-MAV localization is expected to be performed on each MAV individually, whenever a new image is received, whereas inter-MAV localization is only performed in an on-demand fashion. In order to keep communication bandwidth requirements low, the algorithm was designed in a way that it does not require images to be transferred between vehicles whenever inter-MAV localization is performed. Instead, when a client vehicle needs inter-MAV localization data from a host, a packet containing the feature keypoint locations and descriptors from the image obtained by the client is transmitted to the host. The size required per `feature' would be a sum of the size for storing a single keypoint location: two pixel coordinates, thus a combination of two double precision numbers and thus 16 bytes and a 512-bit descriptor vector, so 64 bytes. The total requirement per feature is under 100 bytes, which stays the same for both CPU and GPU implementations of the feature algorithms. For a typical image, the total information that needs to be transferred would be in the order of kilobytes, an amount that can be handled with ease by conventional Wi-Fi networks. After the host vehicles finishes its computation of the client's pose, the pose data transmission involves sending six double precision values: three positions and three orientations, again only about 50 bytes, through a simple $\mathcal{O}(1)$ update (per host). 

Delays in communication were not explicitly considered in the formulation of this problem, but one possible way of adapting to delays can be easily identified. If an inter-MAV localization process is delayed but received at a later time, it is possible to continue localizing using intra-MAV measurements, but when the inter-MAV measurement is received, the system can revert to the previous time, update with the inter-MAV measurement and then re-propagate to the present. While this would involve keeping track of not only the posterior belief of the Kalman filter, but also the measurements: because the measurements are considered to be essentially just the state vector, the space requirements for storing these are significantly lower than those of an implementation that considers map points as part of the state.

In table 3, we show a sample breakdown of how long the various modules that form the VCL algorithm took to execute in our implementation. As mentioned before, in our implementation, the algorithm was run offline: and the data in table 3 was obtained by averaging from a pipeline that performs four functions using data from two vehicles. 

\begin{enumerate}
	\item Detect features in two images (GPU method), construct an initial map.
	\item Perform intra-MAV localization for two new images $I_0$ and $I_1$.
	\item Perform inter-MAV localization between $I_0$ and $I_1$ and fuse with intra estimate computed for $I_1$.
	\item Periodically, detect features in two new images and update existing map.
\end{enumerate}

The average times taken executing these modules is shown in the table. We note here that these numbers are purely for indicative purposes: to help gauge the relative differences between, for example, inter and intra-MAV localizations. The true computational load on any given hardware depends on multiple factors such as existing load on the computer, size of images, number of features being handled during matching/estimation etc. At the same time, the algorithm can also be improved through code optimization, usage of multithreading etc.

\begin{table}
	\centering
	\begin{tabular}{c | c | c}
		\hline
		Procedure  & \makecell{Percentage \\ computation}  & \makecell{Nominal \\ time (ms)}  \\
		\hline
		Map construction & 35 & 31 \\
		Intra-MAV estimation & 17 & 15.6 \\
		Inter-MAV estimation & 19 & 16.8 \\
		Fusion & 2 & 1.8 \\
		Map update & 27 & 24 \\
		\hline
	\end{tabular}
	\caption{Sample computational time required for each module in the VCL algorithm}
\end{table}

\section{Conclusions}

In this paper, we present a collaborative localization framework aimed at vision based micro aerial vehicles: particularly focusing on those equipped with monocular cameras. Feature detection and matching between the MAVs enables the creation of a 3D map that is then shared between them. Once a map is available, the MAVs are capable of alternating between intra-MAV localization: where each MAV attempts to track features from the map and estimate its own pose; and inter-MAV localization, which comes into picture when the feature tracking suffers: where one/more `host' MAVs attempt to correct the pose of a client MAV using a relative pose measurement under the assumption that feature overlap exists between the host(s) and the client. Intra-MAV and inter-MAV measurements can also be combined in a consistent fashion to result in fused localization estimates that were shown to be consistently more accurate than just intra-MAV estimates. Assuming sufficient feature overlap exists, some of the major factors influencing the effect of collaboration in this framework were shown to be the frequency at which relative measurements are obtained, as well as the number of other vehicles contributing to fusion. This algorithm was tested for image datasets coming from the MAV simulator Microsoft AirSim as well as some from real flights, which validate the accuracy of pose estimation as well as the effectiveness of the contribution of multiple MAVs, thus demonstrating that collaboration has a positive effect on localization through reduction of position/orientation estimation error. 

\subsection{Applications and Future Work}

Given the formulation of the localization problem, we identify two applications where this algorithm could be a good fit. Through the collaborative nature of multiple cameras through robust relative and individual pose estimation, the VCL algorithm can be used to create what we refer to as a `decoupled aerial stereo' system. In this system, a combination of small, cheap MAVs equipped with a camera can be converted into a variable, high-baseline stereo imaging system. These MAVs, as they are not restricted in baseline or positioning can be used to image large natural structures, buildings etc. or reconstruction through structure from motion. An application such as this would usually guarantee sufficient feature overlap between the MAVs, thus the inter-MAV localization can be used to its full potential. Another possible application for this algorithm would be in cooperative assembly. In cooperative assembly, multiple small MAVs are required to work in a cooperative fashion to lift, carry and arrange payloads significantly larger than the vehicles themselves. This application requires precise localization, and when multiple MAVs attempt to lift/carry an object, localizing with respect to each other enables more precise positioning. This collaborative nature of localization can continue throughout the flight, thus minimizing the chance of instability or loss of payload through drift. This VCL algorithm also forms the base for our work on collaborative uncertainty-aware path planning for MAVs \cite{8462910}.

There are numerous possibilities for extension and future improvements of this work. We identify that adapting the current software into a real-time framework, capable of running onboard separate vehicles with communication capabilities would be an important extension. Such a deployment could also investigate the ability to handle communication delays, as well as the creation of a feedback control loop to allow for higher level features such as planning and control to take advantage of this collaborative localization scheme. The algorithms used in the VCL framework could also be extended to match a more robust SLAM framework: for instance, generating and including map uncertainty could help in better mapping. It is also possible to use a more accurate system model within the Kalman filter to match the MAV dynamics. Another direction of work could involve integrating IMU measurements into the vision based framework in order to relax the assumption of knowing an initial estimate of scale: utilizing visual-inertial data would allow for online scale estimation and propagation.

\begin{acknowledgements}
  The authors would like to thank Keith Sponsler and Subodh Mishra for their invaluable help with flight testing and data collection.
\end{acknowledgements}

%
%


\end{document}